%% file: main.tex
\documentclass[10pt,twocolumn,letterpaper]{article}

\usepackage[pagenumbers]{cvpr} % To force page numbers, e.g. for an arXiv version

\usepackage{hhline}
\usepackage{makecell}
\usepackage{multirow}
\usepackage{lipsum}
\usepackage{afterpage}
\usepackage{colortbl}
\usepackage[accsupp]{axessibility} % Improves PDF readability for those with visual impairments.

\input{preamble}

\definecolor{cvprblue}{rgb}{0.21,0.49,0.74}
\usepackage[pagebackref,breaklinks,colorlinks,citecolor=cvprblue,]{hyperref}
\def\paperID{6192} % *** Enter the Paper ID here
\def\confName{CVPR}
\def\confYear{2024}

\title{T4P: Test-Time Training of Trajectory Prediction via Masked Autoencoder \\ and Actor-specific Token Memory}

\author{Daehee Park, Jaeseok Jeong, Sung-Hoon Yoon, Jaewoo Jeong, and Kuk-Jin Yoon\\
Visual Intelligence Lab., KAIST, Korea\\
{\tt\small \{bag2824,jason.jeong,yoon307,jeong207,kjyoon\}@kaist.ac.kr}
}

\begin{document}
\maketitle
\input{sec/0_abstract}    
\input{sec/1_intro}
\begin{figure*}[t]
    \centering
    \includegraphics[width=0.99\linewidth]{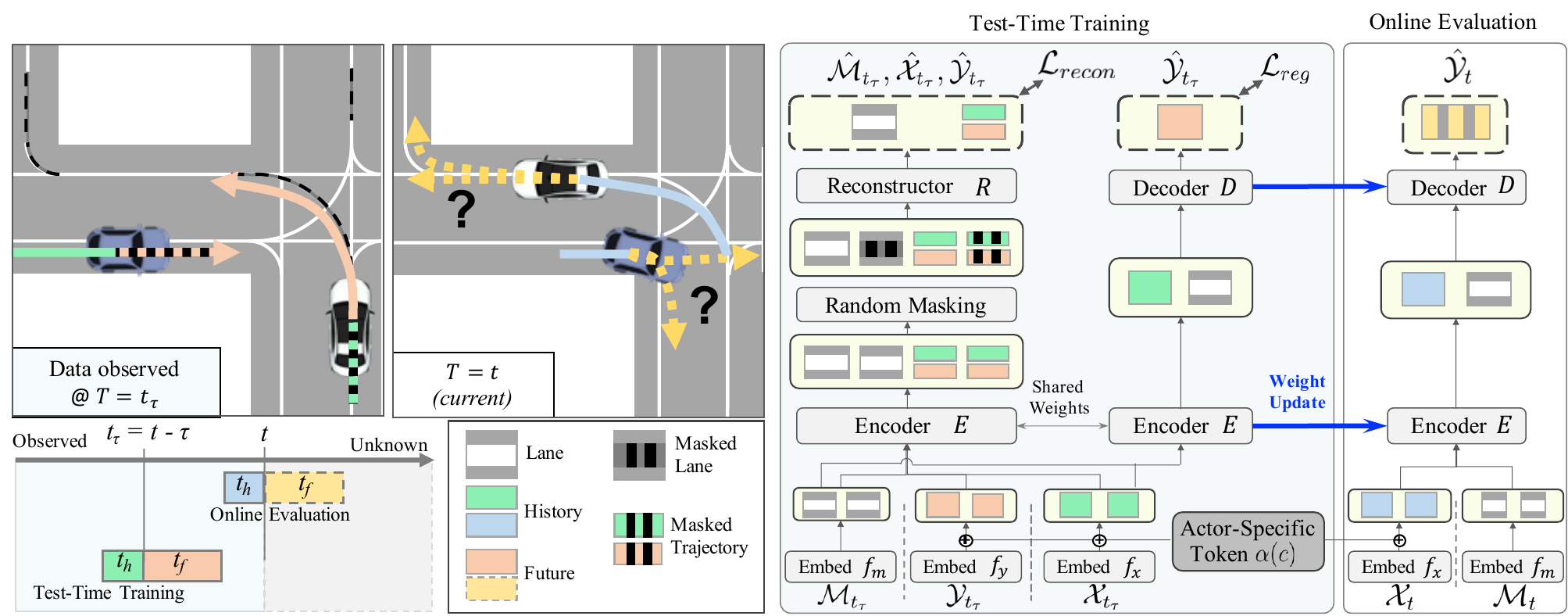}
    \vspace{-5pt}
    \caption{Overall method. During test-time training, the network trained on source dataset is optimized on target data under online setting. The model is optimized both from regression and reconstruction loss. Both losses utilize the data observed at the delayed time stamp ($t_\tau$). Actor-specific token is used to learn instance-wise motion pattern during test-time training phase. During online evaluation phase, model and actor-specific token learned from test-time training phase are used.}
    \label{fig:method_overall_method}
    \vspace{-7pt}
\end{figure*}
\input{sec/2_related}
\input{sec/3_method}
\input{sec/4_experiment}
\input{sec/5_result}

\input{sec/6_ablation}
\input{sec/7_conclusion}

{
    \small
    \bibliographystyle{ieeenat_fullname}
    \bibliography{main}
}


\end{document}

% \maketitle
% \input{sec_template/0_abstract}    
% \input{sec_template/1_intro}
% \input{sec_template/2_formatting}
% \input{sec_template/3_finalcopy}

\input{sec/X_suppl_arxiv}
{
    \small
    % \bibliographystyle{ieeenat_fullname}
    % \bibliography{main}
}

%% file: preamble.tex
\usepackage[dvipsnames]{xcolor}

%% file: sec/0_abstract.tex
\begin{abstract}
% Trajectory prediction은 multi-agent 간의 상호작용과 도로 구조와 같은 주변 환경 고려가 필요한 challenging 한 problem 이다.
% 이 복잡한 문제를 해결하기 위해 data-driven 방식으로 해결하는 방법들이 있었지만, 이러한 방식은 training set 과 다른 환경에서는 reliable 한 예측을 할 수 없다는 단점이 있다.
% 이 문제를 해결하기 위해 domain adaptation 을 이용한 몇몇 방법들이 제안되었지만, 데이터셋에 전혀 포함되어 있지 않은 data가 test에 들어오는 경우에 대해서는 underdiscovered 되어있다.
% 따라서 우리는 Training 시에 전혀 본적이 없는 데이터가 들어왔을 때에도 적응하는 최초의 test time training (ttt) method for trajectory prediction 를 제안한다.
% Trajectory prediction 은 auto-labeling task로, 다른 ttt framework 들과는 달리 학습할 gt가 test 시에도 존재한다.
% 하지만 우리는 trajectory prediction task 에 특화된 두가지 관점에서 이러한 gt 기반 online learning 방법을 보완한다.
% 첫번재, 여러 agent 와 차선같은 주변 환경의 복잡한 상호작용을 잘 모델링 해야한다는 점을 지적한다.
% 이를 위해 test time 에 mae framework 를 이용해 이들간의 상호작용을 잘 capture 할 수 있도록 유도한다.
% 두번째, 주행 패턴이 driver 마다 달라진다는 점을 지적한다.
% 이를 위해 driver-specific token memory bank 를 활용하여, 개인화된 학습가능한 token 을 decoding 시에 활용할 수 있는 구조를 제안한다.
% 제안한 방법은 nuscenes, lyft, waymo, interaction 등 다양한 데이터셋에서 검증되었으며, 기존의 sota ttt 방식을 예측 성능에서 큰 차이로 뛰어넘었으며, 효율성 측면에서도 더 뛰어남을 보였다.
% 아래는 chatgpt 버전

Trajectory prediction is a challenging problem that requires considering interactions among multiple actors and the surrounding environment.
While data-driven approaches have been used to address this complex problem, they suffer from unreliable predictions under distribution shifts during test time.
Accordingly, several online learning methods have been proposed using regression loss from the ground truth of observed data leveraging the auto-labeling nature of trajectory prediction task.
We mainly tackle the following two issues. 
First, previous works underfit and overfit as they only optimize the last layer of motion decoder. 
To this end, we employ the masked autoencoder (MAE) for representation learning to encourage complex interaction modeling in shifted test distribution for updating deeper layers.  
% We tackle two issues of the previous methods.
% First, we tackle that online adaptation is done with a few GT samples in a delayed time stamp.
% It makes underfitting or overfitting, so previous works only optimized the last layers of motion decoder.
% We employ the masked autoencoder (MAE) for representation learning to encourage complex interaction modeling in shifted test distribution for updating deeper layers.
Second, utilizing the sequential nature of driving data, we propose an actor-specific token memory that enables the test-time learning of actor-wise motion characteristics. 
% Second, we emphasize that driving data comes sequentially during test time, unlike random shuffles during training time.
% Therefore, we can access a specific past motion pattern for each actor instance.
% To address this, we propose an actor-specific token memory, enabling the learning of personalized motion pattern during test-time training.
Our proposed method has been validated across various challenging cross-dataset distribution shift scenarios including nuScenes, Lyft, Waymo, and Interaction. 
Our method surpasses the performance of existing state-of-the-art online learning methods in terms of both prediction accuracy and computational efficiency.
The code is available at \url{https://github.com/daeheepark/T4P}.

\end{abstract}

%% file: sec/1_intro.tex
\section{Introduction}
\label{sec:intro}
\begin{figure}[t]
    \centering
    \includegraphics[width=0.9\linewidth]{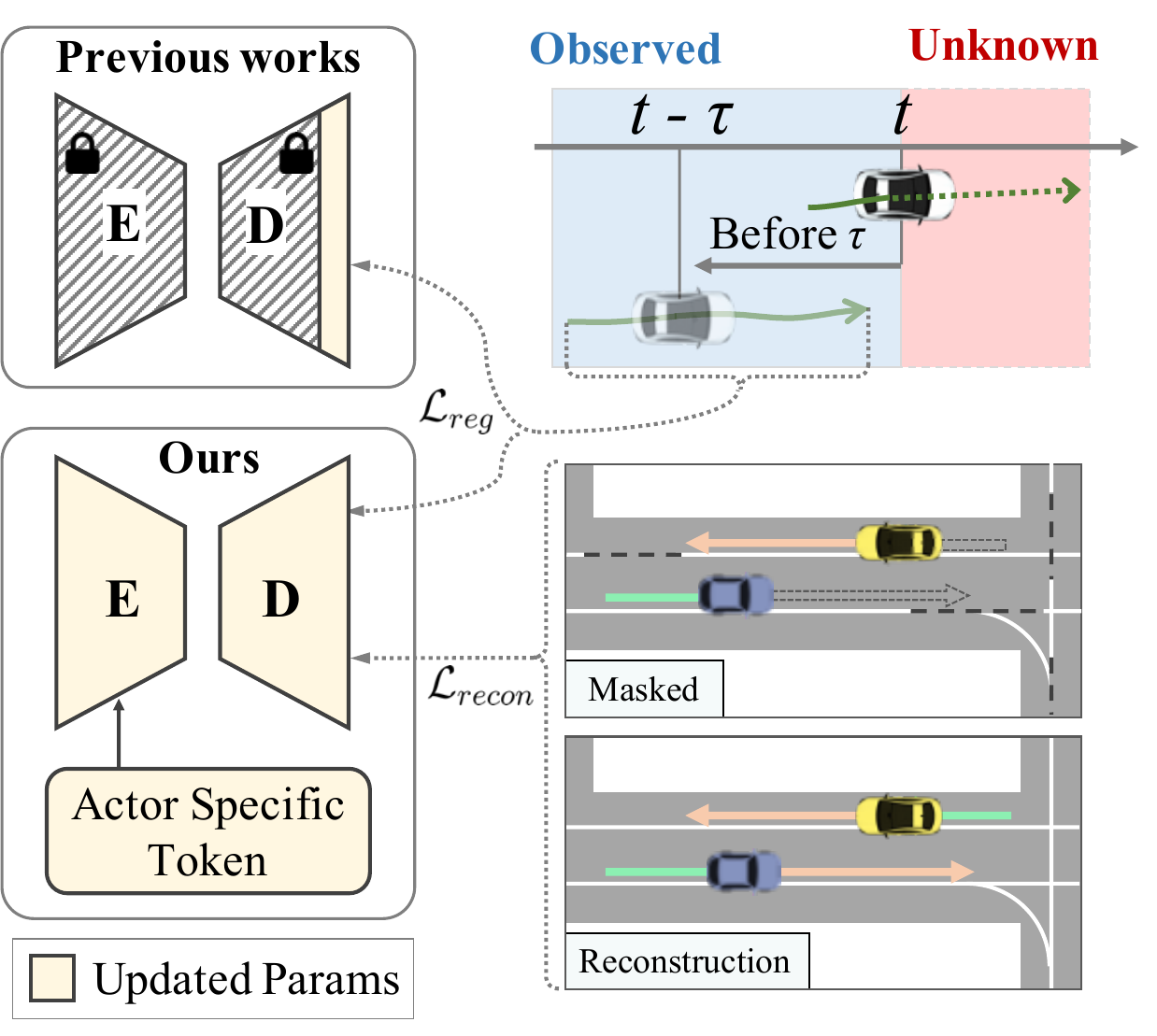}
    \vspace{-5pt}
    \caption{Previous methods optimize the last layer of the decoder using regression loss from delayed ground truth. Our method, on the other hand, learns representation via a masked autoencoder, which boosts prediction performance by optimizing deeper layers. In addition, the proposed actor-specific token enables the prediction model to learn actor-wise motion characteristics.}
    \label{fig:intro}
    \vspace{-12pt}
\end{figure}
Trajectory prediction plays a significant role in autonomous systems, enhancing safety and navigation efficiency~\cite{hu2023planning, gu2023vip3d}. 
Recently, data-driven methods have shown remarkable prediction capabilities~\cite{zhu2023ipcc, xu2023uncovering, zhou2023query, rowe2023fjmp, choi2023r, aydemir2023adapt, jiang2023motiondiffuser, mao2023leapfrog, park2023leveraging, ngiam2022scene}; however, they are prone to distribution shift~\cite{zhang2021deep, chen2023improved}. 
Trajectory prediction models also produce unreliable output when faced shifts in the data distribution~\cite{gilles2022uncertainty}, posing significant risks in various real-world applications.
This vulnerability stems from the ease with which the trajectory data distribution can be altered by numerous factors, such as scene changes and driving habits; \emph{i.e}\onedot road layout, interaction between agents, driver demographics~\cite{liu2022towards, yan2023int2, chandra2019traphic}. 

To address this challenge, recent methods have proposed domain adaptation and generalization strategies which aim to \textit{anticipate} the distribution shifts and accordingly train the model~\cite{xu2022adaptive, huang2022crossdomain, wang2023bridging}. 
However, due to the wide variety of factors influencing data distribution, the anticipated shifts may differ significantly from those encountered at test time.
As a result, several online learning methods have been developed to dynamically adapt models during test time~\cite{wang2022online,ivanovic2023expanding}. 
Since trajectory prediction serves as an auto-labeling task where trajectory data is obtained from object tracking, the observed past and future trajectory provide both input and corresponding ground truth for supervision ($\mathcal{L}_{reg}$), as depicted at the top right of Fig.~\ref{fig:intro}. 
Nevertheless, updating the entire model may ruin the representation learned from the source data, so only part of the network, such as the batch normalization layer, is updated ~\cite{schneider2020improving, yang2022test,kim2022ev,gong2022note}.
Particularly, because regression loss is calculated at delayed timestamps and only a few samples are available during test time, this approach risks deteriorating the model's learned representation~\cite{wang2022online}.
Therefore, previous online prediction methods mainly update only the last layer of the decoder.

In this work, we propose a test-time training (TTT) for trajectory prediction with two key aspects.
First, we build a masked autoencoder (MAE) framework to adapt deep features, incorporating good representation that captures complex interactions between agents and road structures.
% 기존 online learning 방법들이 깊은 레이어을 업데이트 시키지 못하는 것이 학습된 representation 이 망쳐지는것이 원인이므로, 이를 guide 할 수 있는 MAE 를 employ 하였다.
Due to the challenge of existing online learning methods in damaging learned representations when updating deeper layers, we employ a MAE to guide representation learning.
% to incorporate good representation capturing complex inter/intra interaction between agents and road structures. 
% Unlike existing online learning methods that show deteriorating performance while conducting deep layer optimization, MAE guides the network to learn apt feature representations even at the deepest layers. 
Second, we introduce an actor-specific token memory that has significant advantages in real-world driving scenario where data arrives continuously and sequentially.
% Previous works update the model parameters that all actors share.
% However, e
As each actor instance has its own driving habits and the past motion pattern of specific actors can be accessed from the arrived observations, we design a token memory in transformer structure~\cite{vaswani2017attention} and its training strategy to learn actor-wise motion characteristics.
% The actor-specific tokens are initialized from an averaged learned actor class token during the offline training, and instance-wise trained to reflect its characteristics during online training.
The proposed TTT framework is validated on challenging cross-dataset distribution shift cases between nuScenes, Lyft, Waymo, and INTERACTION, and shows state-of-the-art performance surpassing previous online learning methods. 
Furthermore, we show the practicality of our TTT framework by evaluating its computational efficiency.
% Furthermore, we show that our TTT framework can be applied to real-world scenarios by evaluating its computational efficiency.
We summarize our contributions as below:
\begin{itemize}
    \item We propose a test time training for trajectory prediction (T4P) by utilizing a masked autoencoder to learn deep feature representations that stably improve prediction performance across entire network layers.
    \item We introduce an actor-specific token memory used to learn the different actor characteristics and habits.
    \item Our method is validated across 4 different datasets as well as different temporal configurations. Ours achieves state-of-the-art performance both in accuracy and efficiency.
    
\end{itemize}

%% file: sec/2_related.tex
\section{Related Works}
\label{sec:formatting}

\subsection{Trajectory Prediction}
Trajectory prediction garners attention with the emergence of methods that can enhance perception or planning~\cite{li2023modar, chen2023trajectoryformer, chen2022scept}.
Its goal is to predict future trajectories of traffic actors based on their historical trajectories and the context of their environment~\cite{cui2019multimodal, xu2023eqmotion, bae2023eigentrajectory, pourkeshavarz2023learn, jiao2023semi, Zhu_2023_ICCV, shi2023trajectory, dong2023sparse, djuric2020uncertainty}.
Historical trajectories, or tracklets of traffic actors, are sequentially acquired via vehicle detection and tracking systems. 
Some studies employed this temporal property to enhance prediction via memory replay~\cite{li2022graph, rolnick2019experience, isele2018selective}.
In the early stages of trajectory prediction, only the historical trajectory of the actors of interest was considered.
However, recent studies emphasize the significance of understanding interactions among agents~\cite{tsao2022social, yue2022human} and the rules governed by surrounding environments~\cite{wang2022ltp, xu2022pretram} in improving prediction performance.
This has led to the development of models that incorporate multi-head attention or graph-based methods to capture these interactions~\cite{li2022graph, girgis2022latent, gilles2022thomas}. Additionally, MAE has been adopted for pretraining to better understand agent interactions~\cite{cheng2023forecast, chen2023traj}. 
To further refine prediction capabilities, various generative models have been introduced, enabling the generation of future trajectories~\cite{wang2020improving, lee2022muse, xu2022socialvae, choi2022hierarchical}.
% 또한 주변 환경 정보를 잘 인코딩 하기 위해 생성 모델을 활용한 다양한 방법이 제안되었다.
% probabilistic density papers~\cite{sun2023stimulus, chen2023unsupervised, maeda2023fast, }
% many applications~\cite{li2023weakly, rempe2023trace, agro2023implicit}
% minor applications ~\cite{zhang2023trajpac, }

\subsection{Transfer Learning in Trajectory Prediction}
% Data-driven approach 가 superior 한 예측 성능을 내면서, 그에 따라 data-driven approach 의 단점 또한 trajectory prediction 에서 극복해야 할 문제로 떠올랐다.
% Data-driven approach 는 training 시와 test 시 데이터 간에 distribution shift 가 있을 경우 성능이 제한되는 문제점이 있다.
% 이에 따라 몇몇 방법들에서 general domain adaptation 방식을 이용해 distribution shift 를 해결하고자 하였다.
% ~~~ 그런 방법들
% 특히 trajectory prediction 의 특수한 성질을 targetting 하여 domain gap 을 줄이려는 시도가 이었다.
% ~~ 는 도로 구조가 다르다는 점을 착안하여 도로 구조가 달라져도, 데이터 표현 관점에서는 큰 차이가 없도록 하는 ~~를 제안하였다.
% 한편, agent 들간의 interaction,  또는 다양한 측면에서 domain generalization 을 시도한 방법들도 있었다.
% 하지만 이러한 기존 방식들은 모두 domain shift 를 어떻게든 추측하여, 그를 cover 할 수 있는 방법을 제안한 것이다.
% 하지만 앞서 설명했듯, trajectory 데이터는 수많은 요소에 의해 변하기 때문에, 이를 예측하고 대응하는 것은 의미가 없을 수도 있다.
% 따라서 최근에는 online learning 을 이용해 한번도 보지 못한 test set 에 대해 adaptation 을 하는 방식들이 제안되었다.
% 이 방법들은 trajectory prediciton 이 autolabeling task 라는 점을 이용해, test time 때 input 과 gt label 이 tracking history 로써 주어지는 것을 이용했다.
% 하지만 이러한 방법들은 몇개의 sample 로만 network 를 업데이트 시키는 한계로 인해, network 의 마지막 layer 만 adaptation 시키는 것이 그쳤다.
% 실제로, ~ 의 실험에서는 더 깊은 layer 를 학습시키는 것이 예측 성능을 하락한다는 것이 보고되었다.
With data-driven approaches offering superior performance in trajectory prediction, their effectiveness diminishes under distribution shifts~\cite{gilles2022uncertainty, park2023improving}.
In response, several studies have adopted domain adaptation or generalization strategy~\cite{wang2022transferable, wang2022atpfl}. 
Some specifically aimed to reduce the domain gap within unique characteristics of trajectory prediction: differences in road structures~\cite{Ye2023ImprovingTG}, actor interaction~\cite{xu2022adaptive}, \etc.
However, these methods rely on anticipating how to cover domain shifts. 
Yet, given that trajectory data is subject to influence from numerous factors, predicting and accommodating for shifts may not always be sufficient.
Consequently, recent developments have introduced adaptation to unseen test sets using online learning~\cite{li2022online, huynh2020aol, Huynh_2023_WACV}. 
Among them, some methods~\cite{wang2022online,ivanovic2023expanding} showed remarkable prediction performance improvement under severe distribution shifts like cross-dataset cases by utilizing regression loss for online learning.
These methods exploit the fact that the input and ground truth (GT) labels are provided at test time as tracking history. 
However, with the limitations of updating with only a few samples in a delayed time, adaptation becomes restricted to the last layer of the decoder. 
% In fact, ablation experiments have shown that attempting to train deeper layers can lead to a degrading performance.
% There are a line of works that use memory approach for online learning without regression supervision~\cite{Huynh_2023_WACV, }.
% However, they do not target on the case of severe distribution shift, and memory-based methods are known to be limitly applicable to pedestrian, so they are not included for the comparison in this work.

\subsection{Test Time Training}
% Test Time Training (TTT)는 training 때 보지 못했던 Test time 때 들어오는 data가 들어올때, 이 ood data 에 대해서 network 를 training 시키는 것을 의미한다~\ref{liang2023comprehensive}. 
% DG 나 DA 가 training phase 에만 이루어지는 것과는 다르게, TTT는 test phase 에 test data 에 접근할 수 있는 장점을 갖는다.
% test phase 에는 일반적인 경우에는 gt label 을 얻을 수 없기 때문에, TTT 는 주로 unsupervised 또는 self-supervised manner 로 이루어진다.
% unsupervised 방식은 주로 test sample 에 대해 regularization 을 주는 방식으로 네트워크가 adaptation 된다.
% TENT, SHOT, MEMO 은 classification task 에서 Information Entropy 관점에서 regularization 을 수행하였다.
% In line of works, test sample 의 distribution 을 batch normalization statistics 를  Calibration 함으로써 조정하려는 DUA, NOTE, TN-SBI 같은 방법들이 있었다.
% self-supervised manner 방식은 shift 된 data distribution 에서 좋은 representation 을 학습하도록 model 을 optimize 하는 방식이다.
% 그것은 unlabeld data 에서 pretext task 를 통해 supervisory signal 을 이용한다 (OnTA, CluP, SHOT++).
% TTT 는 Y 자 구조의 네트워크를 제안했는데, 각각 feature encoder, selsup branch, decoder branch 로 이루어진 구조이다.
% decoder branch 는 고정시킨 채로, rotation prediction 의 pretext task 를 통해 unseen test image 에서 encoder 와 sepsup branch 를 optimize 함으로써 representation learning 을 하였다.
% rotation prediction을 masek autoencoder framework 로 selsup task 를 대체한 TTT-MAE 도 같은 역할을 수행하며, 이를 point cloud classification 에 활용한 연구도 있었다.
% 우리의 연구는 TTT-MAE 의 학습 framework 를 trajectory prediction 의 특성에 맞도록 gt supervisory signal 에 additional 하게 representation learning 을 수행하는 방식이다.
Test-time training (TTT) is a method that trains the network on test time data, unseen during training~\cite{fleuret2021uncertainty, lim2022ttn, Burns_2021_CVPR, cho2023promptstyler, wang2022continual, chen2022contrastive, tomar2023tesla, ding2022source}. 
Unlike domain generalization or adaptation, which are confined to the training phase, TTT extends model adaptation into the test phase by utilizing available test data~\cite{liang2023comprehensive}. 
TTT methods are categorized into regularization-based approaches for post-hoc regularization of out-of-distribution (OOD) samples~\cite{liang2020we, zhang2022memo}, and self-supervised approaches that employ pretext tasks on test data for optimal representation learning~\cite{li2023robustness, mirza2023mate, chen2023improved, chen2022contrastive, liu2021ttt++}.
Specifically, \textit{TTT}~\cite{pmlr-v119-sun20b} introduced a Y-shaped network structure consisting of a feature encoder, a pretext branch, and a decoder branch. 
The decoder branch is fixed, while the encoder and pretext branch are optimized through self-supervision. 
Adhering to this model, \textit{TTT-MAE}~\cite{gandelsman2022test} integrated a MAE in the pretext task. 
Expanding on this method, we adopt \textit{TTT-MAE} to the domain of trajectory prediction, leveraging its representation learning capabilities to enhance test-time training.
%Our research is a method of performing representation learning additionally with GT supervisory signal on the learning framework of TTT-MAE, which is suitable for the characteristics of trajectory prediction.
% Because complex interaction between agents and agents, or agent and environment is crucial, ou

%% file: sec/3_method.tex
\section{Method}
\subsection{Problem definition}
Trajectory prediction aims to learn the mapping function between the input, consisting of historical trajectory and map information $\textbf{\textup{X}} : \left\{ \mathcal{X}_t, \mathcal{M}_t \right\} $, and the output, consisting of $K$ possible candidates for future trajectory of $N$ traffic actors, $\textbf{\textup{Y}}: \left\{ \mathcal{Y}_t^{0:K-1} \right\}$, at current time $t$.
We predict $C$ different actor classes including vehicle, cyclist, \etc.
Historical and future trajectories are represented as $\mathcal{X}_t = \textbf{\textup{x}}_{t-t_h:t}^{0:N-1}$ and $\mathcal{Y}_t = \textbf{\textup{x}}_{t:t+t_f}^{0:N-1}$ where $t_h$ and $t_f$ represent sequence length of input and output trajectory.
Here, $\textbf{\textup{x}}_t^n$ represents the spatial location of actor $n$ at time $t$.
For map information $\mathcal{M}_t$, we use $L$ segmented lane centerlines around ego-actors
which is widely-used in trajectory prediction methods.

We deal with the case when the target data distribution during test time $\left\{ \textbf{\textup{X}}, \textbf{\textup{Y}} \right\}^T$ is different from the source data distribution seen during the training phase $\left\{ \textbf{\textup{X}}, \textbf{\textup{Y}} \right\}^S$.
We formulate the problem as a online adaptation scenario in which one data sample is given per each time interval as time passes.
The test data is consists of multiple distinct \textit{scenes}.
Each scene includes \textit{temporally ordered} data samples which are captured through real-world driving. 
Following standard TTT methods, there is no access to the source data at test time.
However, thanks to the auto-labeling nature of trajectory prediction, there is access to delayed GT future trajectory ($\textbf{x}_{t_\tau:t_{\tau}+t_f}$) from a previous time window at time $t_\tau (= t - \tau)$ 
as depicted in left lower corner of Fig.~\ref{fig:method_overall_method}.

\subsection{Overall method}
Our method, Test-Time Training of Trajectory Prediction (T4P), enhances the online learning method using supervision from a delayed GT future trajectory with representation learning from MAE and actor-specific token memory.
Following standard \textit{TTT}~\cite{pmlr-v119-sun20b}, the overall framework consists of three phases: \textit{offline training}, \textit{test-time training}, and \textit{online evaluation}. 
Offline training occurs before test time, and test-time training and online evaluation are executed repeatedly and sequentially during test time.
We adopt the ForecastMAE~\cite{cheng2023forecast} backbone consisting of embedding layers $f$, a shared encoder $E$, a reconstruction head $R$ and a motion decoder head $D$, as depicted in the middle of Fig.~\ref{fig:method_overall_method}.
The detailed methods during each phase are described below:

\subsection{Offline training}
During offline training, the model is trained on source data using both reconstruction loss and regression loss.
\begin{equation}
    % \vspace{-5pt}
    \min _{\theta \in \left\{ f, E, R, D \right\} } \mathbb{E}_{ \mathcal{X}, \mathcal{Y}, \mathcal{M} \in \left\{ \textbf{\textup{X}}, \textbf{\textup{Y}} \right\}^S }\left[\mathcal{L}_{recon} + \mathcal{L}_{reg} \right]
    \label{eq:objective_training}
    % \vspace{-0pt}
\end{equation}
In this subsection, subscript $t$ is omitted for simplicity.
First, all input elements ($\mathcal{X}, \mathcal{Y}, \mathcal{M}$) are embedded with their respective embedding layer ($f_x, f_y, f_m$).
Additionally, we define actor class token $\bar{\alpha} \in \mathbb{R}^{C \times D}$ that learns different motion patterns of each actor class.
The actor class token is implemented as a learnable embedding of a transformer structure.
Corresponding actor class token $\alpha(c)$ is added to trajectory embedding according to the class of each actor.
\vspace{-3pt}
\begin{equation}
    h_x, h_y, h_m = f_x(\mathcal{X})+\alpha(c), f_y(\mathcal{Y})+\alpha(c), f_m(\mathcal{M})
\label{eq:training_embedding}
\end{equation}
For reconstruction, the history/future trajectory and lane embeddings are fed to the encoder to obtain the encodings $F_x$, $F_y$, $F_m$.
Then, segments of the encodings are randomly masked and replaced with masking tokens, $M_x$, $M_y$, $M_m$, and the other encodings remain unmasked ($F_x', F_y', F_m'$). 
Here, we use random masking for the lane centerline and complementary masking strategy for history/future trajectory following previous works~\cite{cheng2023forecast}.
The masking tokens and the unmasked encodings are fed to the reconstructor to reconstruct the masked elements.
\begin{equation}
\begin{aligned}
    &F_x, F_y, F_m = E(h_x, h_y, h_m) \\
    &\hat{\mathcal{X}}, \hat{\mathcal{Y}}, \hat{\mathcal{M}} = R( M_x, M_y, M_m, F_x', F_y', F_m' )
\end{aligned}
\end{equation}
The encoder and reconstructor both consist of multi-head attention to utilize interaction between history, future and lanes, thus, the reconstruction guides to the model have the capability of interaction reasoning.
Finally, a reconstruction loss $\mathcal{L}_{recon}$ is computed as the MSE loss between the ground truth and the reconstructed outputs.
\begin{equation}
\small
    \mathcal{L}_{recon} = \frac{1}{N}\sum_{n}( \mathcal{X} - \hat{\mathcal{X}} )^2 + \frac{1}{N}\sum_{n}( \mathcal{Y} - \hat{\mathcal{Y}} )^2 + \frac{1}{L}\sum_{l}( \mathcal{M} - \hat{\mathcal{M}} )^2
\end{equation}

For the decoder head, historical trajectory and lanes embeddings are again fed to the same encoder and the output encodings are passed to the motion decoder, composed of MLP layers.
The decoder outputs K candidates for trajectory prediction, and the regression loss $\mathcal{L}_{reg}$ is computed with the widely-used Winner-takes-all (WTA) loss~\cite{guzman2012multiple, liang2020learning}.
\begin{equation}
\begin{aligned}
    &\hat{\mathcal{Y}}^{0:K-1} = D( E( h_h, h_l ) ) \\
    &\mathcal{L}_{reg} = \frac{1}{N}\sum_{n} \operatorname*{argmin}_{k \in K}  ( \mathcal{Y}^n - \hat{\mathcal{Y}}^{n, k} )^2
\end{aligned}
\end{equation}

\afterpage{
\begin{table*}[t]
\centering
\caption{Adaptation results in various distribution shifts. The model is trained on source dataset, and test-time trained and evaluated on target dataset (\textit{Source} → \textit{Target}). All metrics are better in lower value. The best and second-best results are marked in \textbf{bold} and \underline{underline}.}
\vspace{-5pt}
\resizebox{\textwidth}{!}{
\begin{tabular}{l|ccc|c|ccc|c}
\toprule
\multicolumn{1}{c|}{\multirow{2}{*}{\begin{tabular}[c]{@{}c@{}}mADE$_6$\\ / mFDE$_6$\end{tabular}}} & \multicolumn{4}{c|}{Short-term exp (1/3/0.1)}                  & \multicolumn{4}{c}{Long-term exp (2/6/0.5)}                   \\ \cline{2-9} 
\multicolumn{1}{c|}{}                                                                         & INTER → nuS   & INTER → Lyft  & nuS → Way     & Mean & nuS → Lyft    & Way → Lyft    & Way → nuS     & Mean \\ \hline
Source Only                                                                                   & 1.047 / 2.247 & 1.391 / 2.945 & 0.431 / \underline{1.031} & 0.956 / 2.074 & 1.122 / 2.577 & 0.621 / \underline{1.347} & 1.153 / 2.220 & 0.965 / 2.048 \\
Joint Training                                                                                & 1.116 / 2.445 & 1.553 / 3.458 & 0.472 / 1.125 & 1.047 / 2.343 & 1.108 / 2.597 & 0.638 / 1.404 & 1.091 / 2.031 & 0.946 / 2.011 \\
DUA                                                                                           & 1.118 / 2.455 & 1.516 / 3.352 & 0.516 / 1.294 & 1.050 / 2.367 & 1.365 / 3.257 & 0.790 / 1.868 & 1.270 / 2.634 & 1.142 / 2.586 \\
TENT (w/ sup)                                                                                 & 1.102 / 2.423 & 1.519 / 3.405 & 0.448 / 1.071 & 1.023 / 2.300 & 1.068 / 2.514 & 0.628 / 1.381 & \underline{1.077} / \underline{2.012} & 0.924 / 1.969 \\
MEK ($\tau=t_f/2$)                                                                            & 1.012 / 2.445 & 1.283 / 3.458 & 0.445 / 1.125 & 0.913 / 2.343 & 1.079 / 2.597 & 0.629 / 1.404 & 1.079 / 2.031 & 0.929 / 2.011 \\
MEK ($\tau=t_f$)                                                                              & \underline{0.892} / \underline{1.952} & \underline{0.746} / \underline{1.654} & \underline{0.405} / 1.061 & \underline{0.691} / \underline{1.556} & \underline{1.006} / \underline{2.369} & \underline{0.615} / 1.351 & 1.117 / 2.140 & \underline{0.913} / \underline{1.953}    \\
AML (\textit{K$_0$})                                                                          & 2.093 / 4.697 & 2.695 / 6.677 & 1.624 / 2.139 & 2.137 / 4.504 & 1.787 / 3.067 & 1.322 / 2.571 & 1.618 / 2.999 & 1.576 / 2.879 \\
AML (\textit{full})                                                                           & 1.149 / 2.550 & 1.042 / 2.616 & 0.764 / 1.791 & 0.985 / 2.319 & 1.462 / 2.573 & 0.977 / 2.184 & 1.495 / 2.978 & 1.311 / 2.578 \\

\rowcolor[rgb]{0.9,0.9,0.9} Ours (T4P)                                                                                         & \textbf{0.537} / \textbf{1.137} & \textbf{0.391} / \textbf{0.824} & \textbf{0.336} / \textbf{0.807} &  \textbf{0.421} / \textbf{0.923}   & \textbf{0.776} / \textbf{1.820} & \textbf{0.549} / \textbf{1.171} & \textbf{0.996} / \textbf{1.784} &  \textbf{0.774} / \textbf{1.592}   \\ \bottomrule
\end{tabular}
}
\vspace{-5pt}
\label{tab:result_main_adaptation}
\end{table*}
}

\subsection{Test-time training}
During test-time, a data sample consisting of trajectories and maps arrives sequentially.
Therefore, even though we cannot access the GT future trajectory of current time ($\mathcal{Y}_t$), we can access both the inputs and GT  ($\mathcal{X}_{t_\tau}, \mathcal{Y}_{t_\tau}, \mathcal{M}_{t_\tau}$) at a previous time $t_\tau$. 
With this data, the model is optimized with the same objective as Eq.~\ref{eq:objective_training} with target data distribution instead of source data distribution.
\begin{equation}
    \small
    \min _{\theta \in \left\{ f, E, R, D \right\} } \mathbb{E}_{ \mathcal{X}_{t_\tau}, \mathcal{Y}_{t_\tau}, \mathcal{M}_{t_\tau} \in \left\{ \textbf{\textup{X}}, \textbf{\textup{Y}} \right\}^T }\left[\mathcal{L}_{recon} + \mathcal{L}_{reg} \right]
    \label{eq:objective_test_time_training}
\end{equation}
Unlike existing online learning methods that only utilize regression loss, we incorporate an additional reconstruction loss.
This enables the model to learn a good representation that considers the complex actor-actor and actor-lane interaction even in the unseen target data distribution.
An advantage of representation learning is that the performance stably improves even when the deeper layers are optimized.

\subsubsection{Actor-specific token memory}
\begin{figure}[t]
    \centering
    \includegraphics[width=\linewidth]{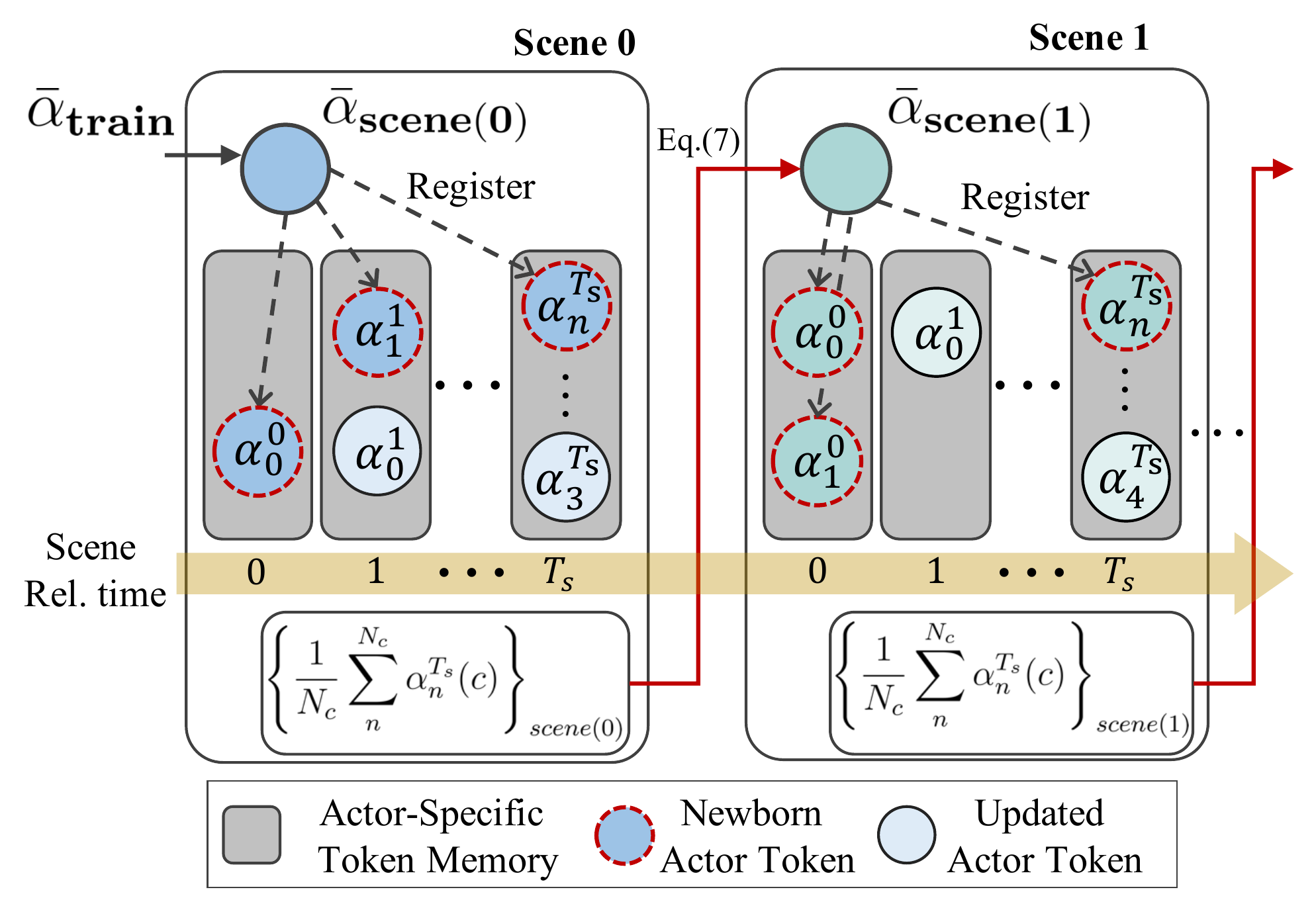}
    \vspace{-18pt}
    \caption{Actor-specific token memory is colored in gray. It evolves as time passes within a scene. For newborn actors, the corresponding class token is registered. Until the actor disappears, the token is updated through test-time training. At the end of the scene, all tokens are averaged by each class and passed to the next scene as denoted in red arrow and Eq.~\ref{eq:scene_change}. }
    \label{fig:method_actor_specific}
    \vspace{-12pt}
\end{figure}
Unlike during the offline training phase, when the data order is shuffled, data at test-time comes in sequentially.
Therefore, at test time, it is possible to keep track of movement patterns of a specific actor instance.
Using this, we propose an actor-specific token memory.

The overall scheme is described in Fig.~\ref{fig:method_actor_specific}.
During offline training, actor class tokens $\bar{\alpha}_{train} \in \mathbb{R}^{C \times D}$ are trained to reflect the average motion pattern of each of the $C$ classes.
At the beginning of the test-time, scene 0, class tokens are initialized from that of training phase ($\bar{\alpha}_{scene(0)} \leftarrow \bar{\alpha}_{train}$).
When a new $n^{th}$-actor appears at time $t$, the class token $\alpha_n^{t} (c)$ is cloned from $\bar{\alpha}_{scene(0)}$ by selecting corresponding class.
The newborn tokens are then registered to the actor-specific token memory.
The token memory is structured as a dictionary where actor instance ID/corresponding tokens are key/values.
At each iteration, the actor-specific tokens are used for both test-time training and online evaluation.
As time progresses, the actor-specific tokens evolve and are updated through the reconstruction and regression losses until the actor disappears in the scene.
By giving each actor its own specific token that distinguishes it from the sharing of other parts of models with other actors, actor-specific motion patterns can be learned.

When the scene changes, scene 1, the actors observed during scene 0 are not to be observed anymore, so we need a new averaged actor class token $\bar{\alpha}_{scene(1)}$.
For that, we average all the tokens in the memory at the final time step $T_S$ of scene 0 after gathering by classes as Eq.~\ref{eq:scene_change}.
Here, $N_c$ refers to the number of actors of class $c$.
It is because, with a sufficient number of actors class tokens per each class in memory, their average motion can be a representative motion pattern of actor classes.
This is more useful than $\bar{\alpha}_{train}$ because newly averaged tokens are trained on the target dataset while $\bar{\alpha}_{train}$ contains motion pattern trained on the source dataset.
The averaged class tokens are then passed to the next scene and used to initialize tokens for the newborn actors.
Please note that while scene 0 is initialized by actor class tokens from the training phase, subsequent scenes obtain as in Eq.~\ref{eq:scene_change}.
More details of memory evolving strategies can be found on the supplementary material.
\begin{equation}
    \bar{\alpha}_{scene (i+1)} \leftarrow \left\{ \frac{1}{N_c}\sum_{n}^{N_c} \alpha_{n}^{T} (c) \right\}_{scene (i)}
\label{eq:scene_change}
\end{equation}

\subsection{Online evaluation}
Using the updated model weight and actor-specific token memory during test-time training, online evaluation is executed.
With the input data ($\mathcal{X}_{t}, \mathcal{M}_{t}$) at current time $t$, the learned encoder and decoder predict multi-modal trajectory ($\mathcal{Y}_t$) for all actors in the sample.

%% file: sec/4_experiment.tex
\section{Experiment}
\subsection{Datasets}
We conducted experiments on well-known datasets, nuScenes~\cite{caesar2020nuscenes}, Lyft~\cite{houston2021one}, WOMD~\cite{ettinger2021large}, and INTERACTION~\cite{zhan2019interaction}, to evaluate T4P on various data distribution shifts.
These datasets are parsed into the same format using \textit{trajdata}~\cite{ivanovic2023trajdata}.
Additionally, to verify in various prediction configurations, experiments were conducted with the two most widely used configurations of long-term and short-term prediction.
Long-term prediction requires predicting 6 seconds into the future given 2 seconds of the past with a time interval of 0.5s, making the input/output sequence lengths to be 5 and 12, respectively.
Short-term prediction requires predicting 3 seconds into the future given 0.9 seconds of the past with a time interval of 0.1s, making the input/output sequence lengths to be 10 and 30, respectively.

\subsection{Implementation details}
For actor classes, we use the 5 classes: \textit{unknown}, \textit{vehicle}, \textit{pedestrian}, \textit{bicycle} and \textit{motocycle}.
Our method predicts $K$=6 future candidates for all actors in the sample.
We use $\tau$ as $t_f$ to enable the past GT future to contain full prediction horizon.
We train and evaluate our model with a single NVIDIA A6000.
Learning rates of model weight and actor-specific parameters are set as 0.01 and 0.5, respectively, and weight decay is set to 0.001 for all.
The gradient is clipped by 15.
For metrics, widely used mADE$_6$ and mFDE$_6$ are used.
Detailed metric definition, model architecture, and training details are included in the supplementary material.

\subsection{Baselines}
We compare our \textit{T4P} with several baselines, including unsupervised/supervised test-time-training methods and online learning trajectory prediction methods.
All baseline methods are implemented using the same backbone.

\noindent \textbf{Source only} refers to the backbone model trained on the source dataset only using regression loss.

\noindent \textbf{Joint training} is similar to source only but trained with regression and reconstruction loss jointly.

\noindent \textbf{DUA}~\cite{mirza2022norm} is an unsupervised post-hoc regularization method only updates batch normalization statistics in a momentum-updating manner without back-propagation.

\noindent \textbf{TENT with supervision} is a variant of the original TENT~\cite{wang2021tent} in which regression loss is used to optimize the batch normalization layers instead of entropy minimization loss, as entropy minimization is not applicable.

\noindent \textbf{MEK}~\cite{wang2022online} is an online learning trajectory prediction method utilizing the Modified Extended Kalman filter.
It uses only regression loss
to optimize the last layer of the decoder.
As the prediction horizon is different in our experiment from the original paper, we use both $\frac{1}{2}t_f$ and $t_f$.

\noindent \textbf{AML}~\cite{ivanovic2023expanding} is an Adaptive Meta-learning method.
Unlike the other methods that use the same backbone, AML replaces the last decoder layer with a Bayesian linear regression layer for adaptive training.
The modified version of backbone without adaptive training is denoted as \textit{K$_0$}, while the full version with adaptive training is denoted as \textit{full}.

\afterpage{
\begin{figure}[t]
    \centering
    \includegraphics[width=0.8\linewidth]{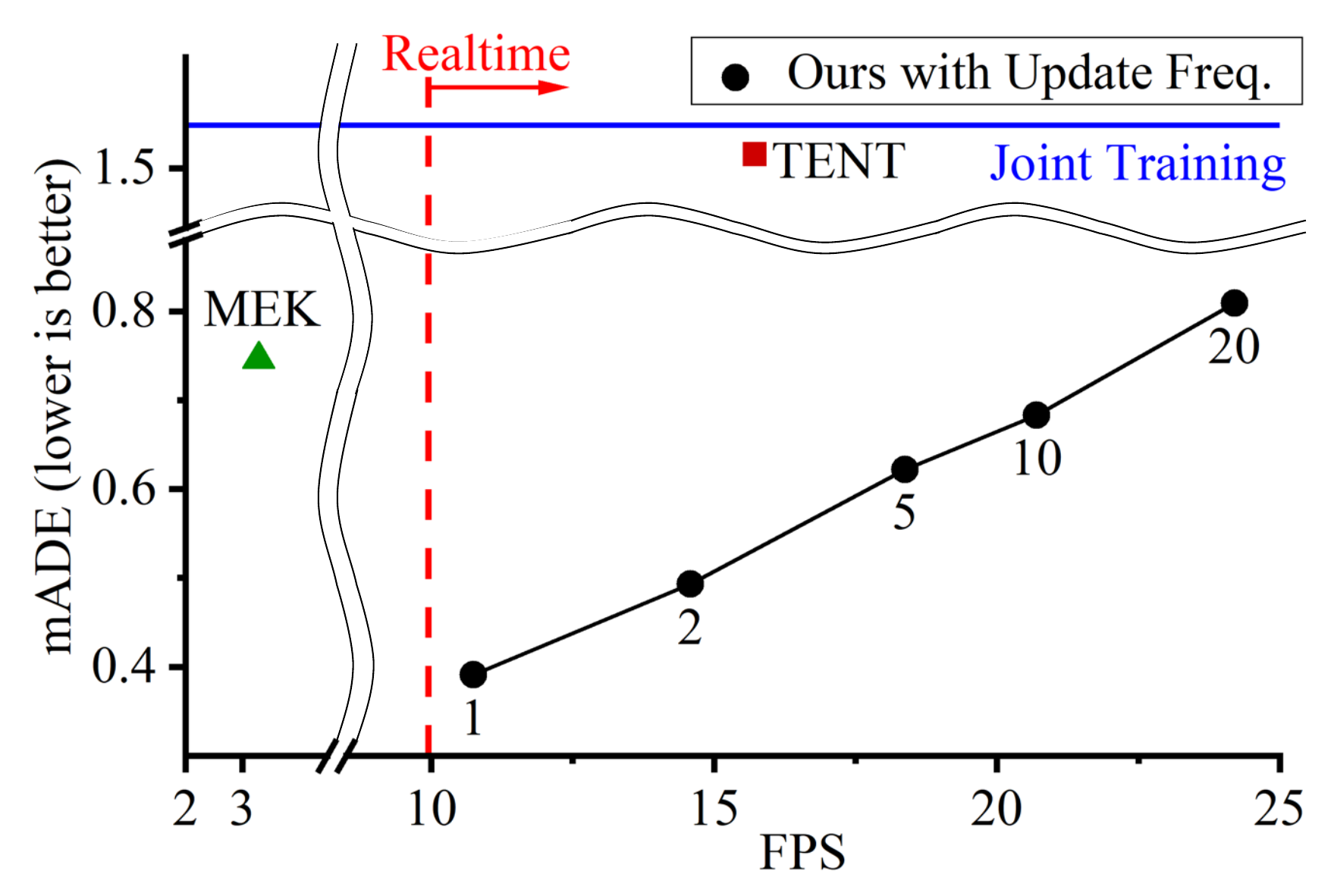}
    \vspace{-8pt}
    \caption{Prediction accuracy and execution time on INTER → nuS (1/3/0.1) experiment. Adjusting update frequency can balance between accuracy and efficiency. Our method significantly outperforms the baseline methods in both accuracy and efficiency.}
    \label{fig:result_exec_time_short}
    \vspace{-8pt}
\end{figure}
}

%% file: sec/5_result.tex
\begin{figure*}[t]
    \centering
    \includegraphics[width=0.99\linewidth]{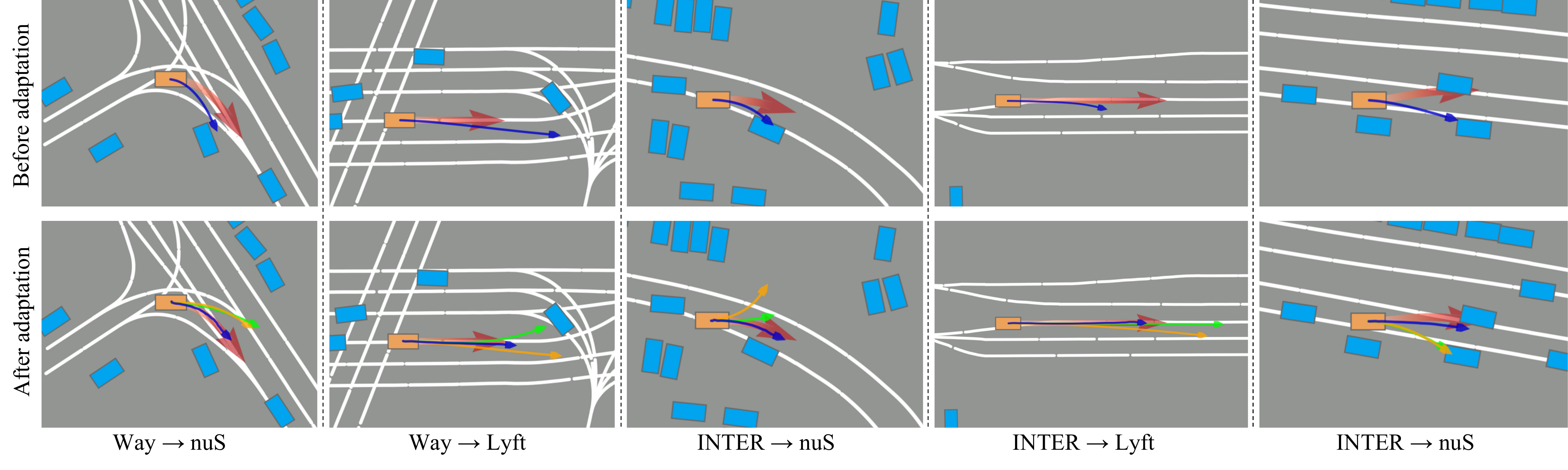}
    \vspace{-10pt}
    \caption{The first row shows prediction before adaptation, and the second row indicates adaptation results by three methods: ours (blue), TENT w/ sup (orange) and MEK (green). Sky blue and orange boxes refer to surrounding actors and actors to be predicted. We depicted only one actor result and one mode among multi-modal predictions closest to the GT for visual simplicity. Please note that our method is multi-modal prediction for all actors method.}
    \label{fig:qualitative_compare}
    \vspace{-10pt}
\end{figure*}
\section{Results}
\subsection{Quantitative results}
The results of comparing our method with the baselines in various distribution shift scenarios are presented in Tab.~\ref{tab:result_main_adaptation}.
We reported three distribution shift scenarios per each time configuration on the table, and other results are included in the supplementary material.
Notably, our approach consistently surpassed baseline performance across all scenarios.

DUA consistently exhibits compromised performance across nearly all cases, a consequence of the distinctive features inherent to the trajectory prediction task.
In contrast to tasks like classification, where a data is treated as a singular sample, trajectory prediction involves multiple agents, each exhibiting distinct motion patterns, within a single data. 
Consequently, holistically updating batch statistics proves to be counterproductive.
Similar challenges are encountered by TENT w/ sup. 
While regression loss prevents a decline in performance, updating only the batch norm layer has little to no effect on the prediction performance. 

MEK exhibited the most substantial performance among the baselines.
While prediction performance improved significantly in short-term settings, the performance showed limited improvement in long-term scenarios. As the Kalman filter updates based on the number of prediction steps, short-term configurations with 12 update steps shows a better performance improvement than long-term configurations with only 5 update steps.

Although AML led to a considerable improvement in the \textit{full} version, the predictive performance itself was substantially degraded due to the significant performance drop in the modified backbone (\textit{K$_0$}).
The limitation of the backbone is due to the Bayesian regression layer being based on probability sampling which is known to be worse than non-probability sampling method of ours~\cite{bae2022non}.
In contrast to all the baseline methods, our method demonstrated state-of-the-art performance in all scenarios featuring various distribution shifts, whether short-term or long-term, showcasing the generalizability of our approach.

\subsubsection{Efficiency}
As efficiency is a crucial factor in TTT, we evaluate the frame per second (FPS) along with accuracy (mADE$_6$), shown in Fig.~\ref{fig:result_exec_time_short}.
We set the performance of the joint training method w/o adaptation as the benchmark and present MEK and TENT, which demonstrates competitive performance among the baselines.
Our approach allows for the adjustment of the update frequency, with a frequency of 1 indicates updating at every opportunity, and 2 means updating every other opportunity.
While frequent updates improve prediction performance, they also increase execution time; adjusting the update frequency allows for a balance between efficiency and accuracy.
As shown in Fig.~\ref{fig:result_exec_time_short}, our method outperforms in both accuracy and efficiency.
Given that the time interval is 0.1 seconds, real-time execution requires a processing speed of at least 10 FPS.
Even at the maximum update frequency of 1, our method maintains real-time capability with a superior accuracy of 0.39.
When increasing the update frequency to 20, the error increases to 0.81, close to MEK's 0.75. 
However, the FPS reaches 24.2, demonstrating overwhelmingly faster operation compared to MEK's speed of 3.3 FPS.
\begin{figure}[t]
    \centering
    \includegraphics[width=0.99\linewidth]{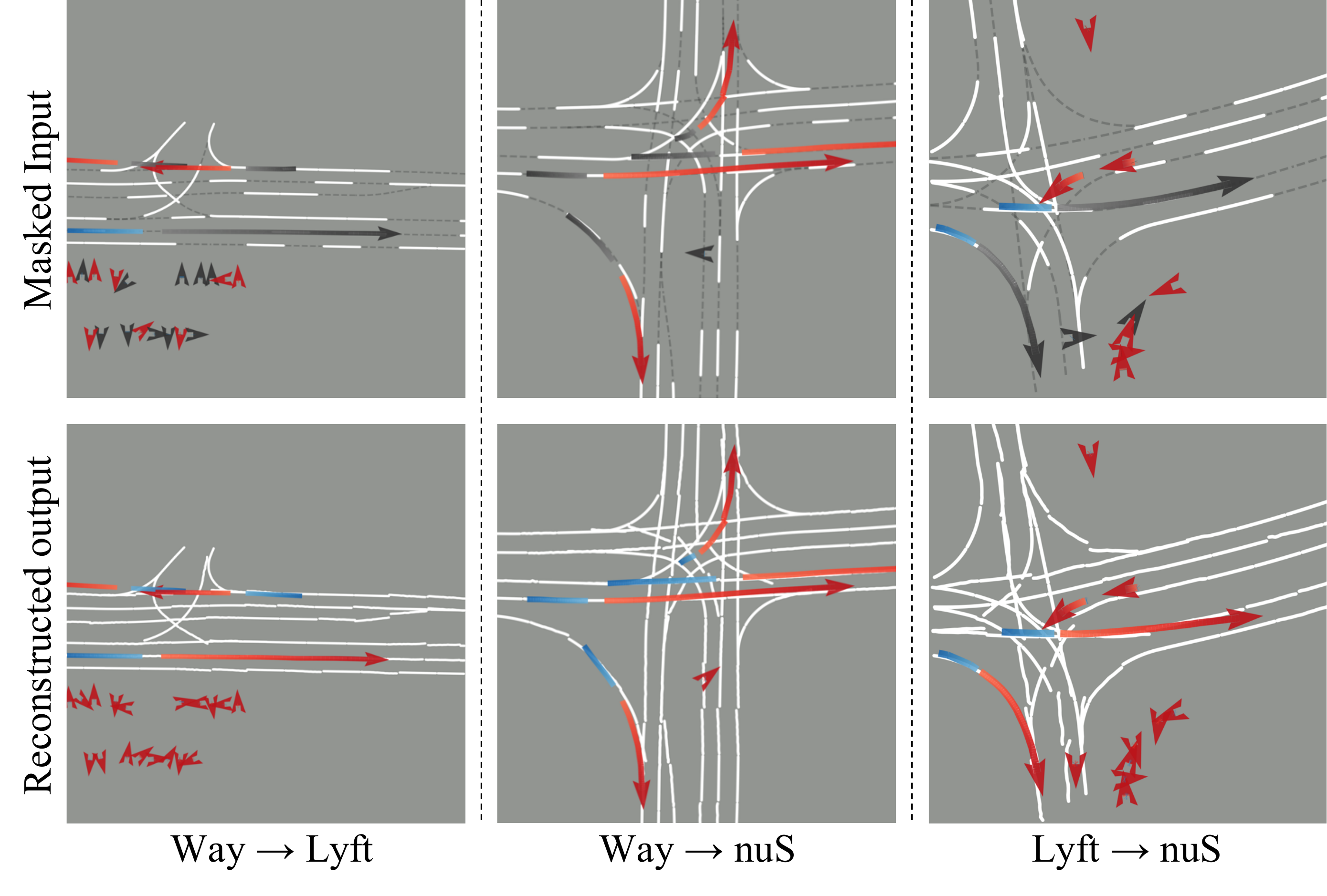}
    \vspace{-10pt}
    \caption{
    %Learned reconstruction result during test-time. 
    The first row indicates masked samples, and the row below shows the reconstructed outputs. The blue/red arrows indicate historical/future trajectories. The black arrows refer to the masked trajectories. The white lines are the lane centerlines, and the gray dashed lines are the masked lane centerlines.}
    \label{fig:qualitative_mae}
    \vspace{-5pt}
\end{figure}
\begin{figure}[t]
    \centering
    \includegraphics[width=0.99\linewidth]{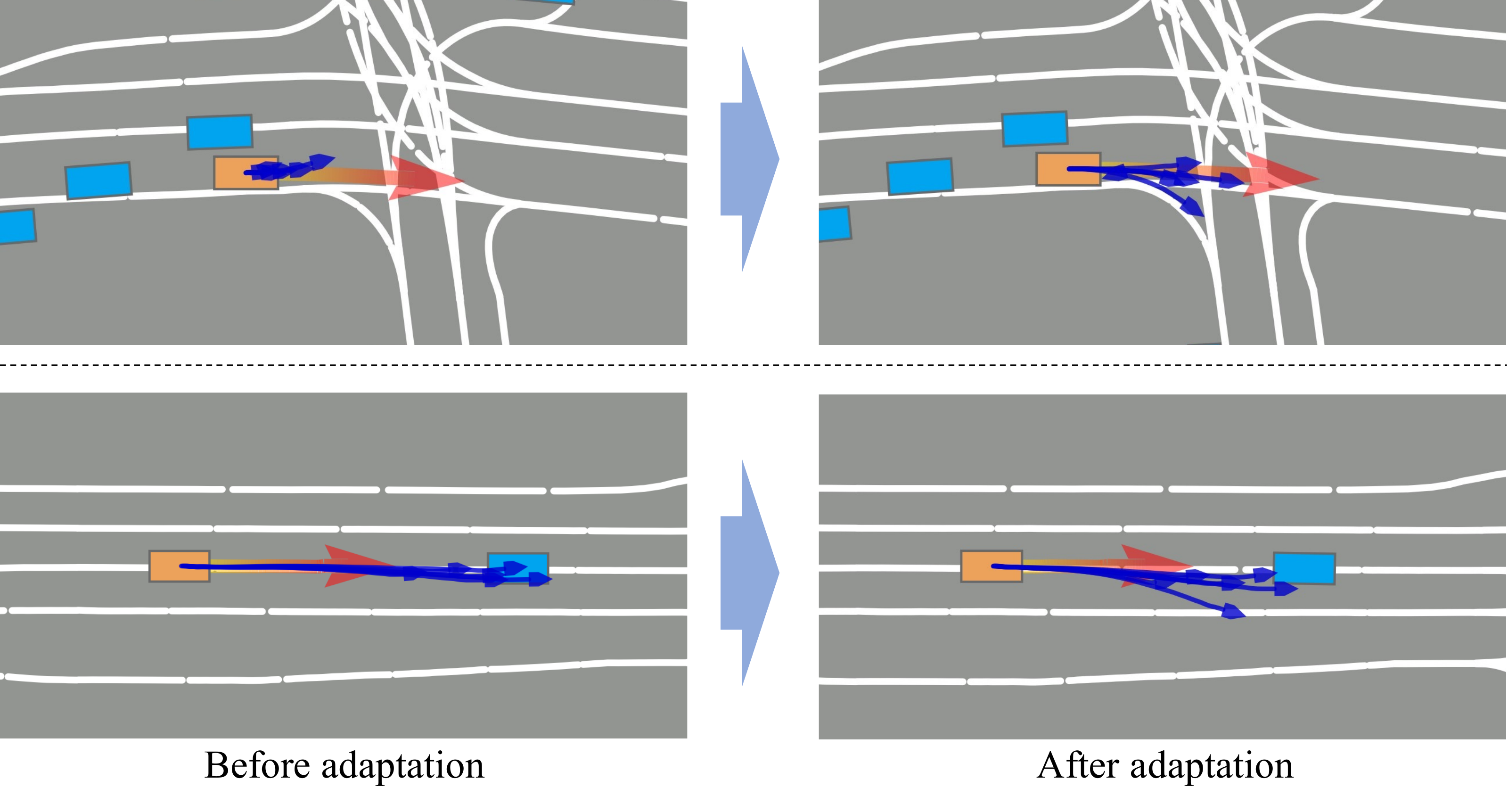}
    \vspace{-10pt}
    \caption{Multi modal prediction results (blue arrow) before and after adaptation via our method. Ours generates elaborate samples that consider interaction between lane (above) or other actor (below) due to representation learning, which cannot be learned from the GT (red arrow) via regression loss.}
    \label{fig:qualitative_multimodal}
    \vspace{-10pt}
\end{figure}
\subsection{Qualitative results}
\noindent \textbf{Comparison to the baselines}: 
We compare our results with TENT and MEK in Fig.~\ref{fig:qualitative_compare}.
While all methods, including ours, perform multi-agent, multi-modal predictions, we only illustrate one actor and the closest mode to the GT for visual simplicity.
The first row of the figure represents predictions before adaptation, and below are the results after adaptation using three different methods.
MEK and TENT exhibit instances of underfitting or excessive overfitting upon adaptation, whereas our method consistently demonstrates stable and accurate predictions. 

\noindent \textbf{Reconstruction results}: 
Reconstruction examples are depicted in Fig.~\ref{fig:qualitative_mae}.
For all agents and lanes within a data sample, random masking is applied, as shown in the first row.
During test-time training, learning for reconstruction is conducted, resulting in successful reconstruction for data with different distributions, as seen in the second row.

\noindent \textbf{Multi-modal prediction results}:
As multi-modality is a crucial issue~\cite{sun2021three, chen2022multimodal, shi2023representing}, we show that ours can handle multi-modal prediction results in Fig.~\ref{fig:qualitative_multimodal}.
The adapted predictions showcase diverse yet plausible scenarios, either considering the lane structure (above) or surrounding agents (below). 
These elaborated samples, although not present in the observed ground truth (GT) future, are learned through the representation learning from reconstruction loss. 
In addition, it shows that actor-specific tokens do not induce mode collapse to only one motion. 
% This is achieved by using the WTA loss and non-probabilistic sampling. 

%% file: sec/6_ablation.tex
\section{Ablation}
\begin{table}[t]
\centering
\small
\caption{Effect of type of losses to be optimized. 
Optimizing all losses shows optimal test-time adaptation performance.
% in long/short-term experiments.
}
\vspace{-5pt}
\label{tab:ablation_loss_type}
\begin{tabular}{c|ccc|c}
\toprule
                                                                                      \multirow{2}{*}{Exp.}  & \multicolumn{3}{c|}{Loss type}                      & \multirow{2}{*}{\begin{tabular}[c]{@{}c@{}}mADE$_6$\\ /mFDE$_6$\end{tabular}} \\ \cline{2-4}
                                                                                        & \makecell{Actor \\ recon} & \makecell{Lane \\ recon} & Reg &                              \\ \hline
\multirow{5}{*}{\begin{tabular}[c]{@{}c@{}}\makecell{INTER $\rightarrow$ Lyft\\ (1/3/0.1)}\end{tabular}} &                  &                 &               & 1.553 / 3.458                \\
                                                                                        & \checkmark                &                 &               & 1.054 / 2.007                \\
                                                                                        & \checkmark                & \checkmark               &               & 0.842 / 1.512                \\
                                                                                        &                  &                 & \checkmark             & 0.674 / 1.430                \\
                                                                                        & \checkmark                & \checkmark               & \checkmark             & \textbf{0.391 / 0.824}       \\ \hline
\multirow{5}{*}{\begin{tabular}[c]{@{}c@{}}\makecell{nuS $\rightarrow$ Lyft\\ (2/6/0.5)}\end{tabular}}    &                  &                 &               & 1.108 / 2.597                \\
                                                                                        & \checkmark                &                 &               & 0.987 / 2.304                \\
                                                                                        & \checkmark                & \checkmark               &               & 0.973 / 2.280                \\
                                                                                        &                  &                 & \checkmark             & 0.942 / 2.262                \\
                                                                                        & \checkmark                & \checkmark               & \checkmark             & \textbf{0.776 / 1.820}      \\ \bottomrule
\end{tabular}
\vspace{-5pt}
\end{table}
\begin{table}[t]
\footnotesize
\caption{Ablation on the depth of optimizing layer according to the loss types. 
$D$, $E$, and $f$ represent the Decoder, Encoder, and Embedding layers, respectively.
The right side of the table indicates the optimization of deeper layers.
Using regression loss only deteriorates performance when optimizing all layers while ours stably and increasingly improves as deeper. }
\vspace{-5pt}
\label{tab:ablation_update_layer}
\centering
\begin{tabular}{l|ccc}
\toprule
           \multirow{2}{*}{Loss} & \multicolumn{3}{c}{Optimizing layers}               \\
        & $D$             & $D$+$E$           & $D$+$E$+$f_{h,f,l}$ \\ \hline
$\mathcal{L}_{reg}$    & 0.864 / 2.072 & \textbf{0.840} / \textbf{2.093} & 0.942 / 2.262       \\
$\mathcal{L}_{reg}+\mathcal{L}_{recon}$ & 0.859 / 2.060 & 0.813 / 1.923 & \textbf{0.776} / \textbf{1.820} \\ \bottomrule     
\end{tabular}
\vspace{-10pt}
\end{table}
\subsection{Reconstruction objective}
Table.~\ref{tab:ablation_loss_type} shows ablation studies on optimizing different loss types.
Both reconstruction and regression losses individually boost prediction performance, with their joint optimization yielding even greater improvements.
Table.~\ref{tab:ablation_update_layer} compares the effects of using only regression loss versus both losses on prediction performance across different layer depths.
% presents the prediction performance according to the depth of the layer updated, comparing the effects of using only regression loss against two losses.
Updating just the decoder ($D$) shows similar results in both scenarios, but extending updates to the encoder ($E$) significantly enhances performance when using both losses.
Furthermore, extending updates to the embedding layers ($f_{h,f,l}$) deteriorates performance when only regression loss is optimized.
This highlights the importance of incorporating representation learning through the MAE, as relying solely on regression loss can lead to suboptimal adaptation and damage to learned representations.
 
% This underscores that without conducting representation learning through the MAE, regression loss alone may suffer from unoptimal adaptation results, leading to the loss of learned representations from the source data.

\begin{figure}[t]
    \centering
    \includegraphics[width=0.99\linewidth]{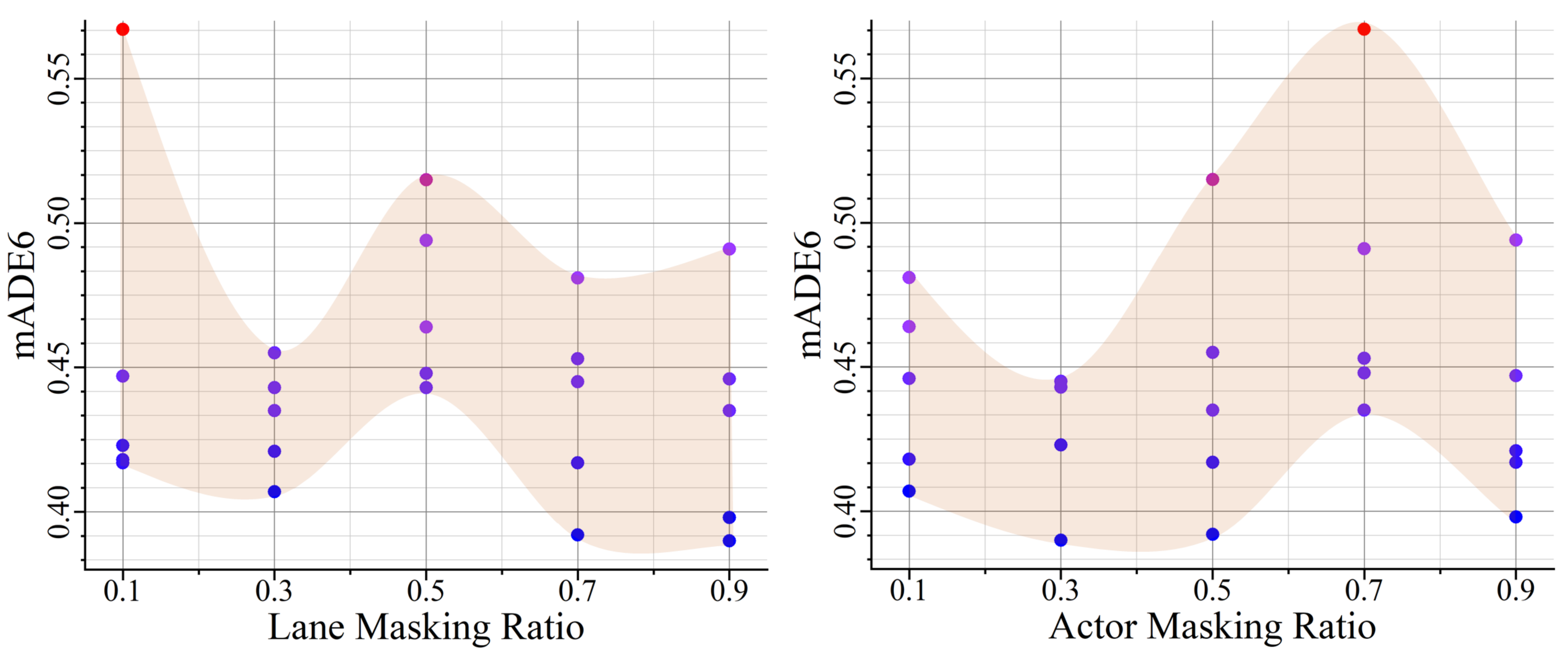}
    \vspace{-3pt}\caption{Tendency of mADE$_6$ according to actor and lane masking ratio respectively. (INTER $\rightarrow$ Lyft (1/3/0.1))}
    \label{fig:ablation_mask_graph}
    \vspace{-5pt}
\end{figure}

We also conduct ablation studies on the masking ratio for both actors and lane centerlines in Tab.~\ref{tab:ablation_mask_table}.
The result is visualized via graphs in Fig.~\ref{fig:ablation_mask_graph} according to lane masking ratio and actor masking ratio, respectively. 
Around 0.3 of lane and 0.4 of actor masking ratio, tendencies of mADE$_6$ follow a U-shape.
In case of too small masking ratio, the reconstruction does not learn sufficient representation from the loss, while large masking ratio interrupt interaction learning due to absence of sufficient information.
However, in both lane and actor masking, mADE$_6$ gets improved when the masking ratio increases above 0.8.
In that case, the reconstruction network is induced to learn scene-specific information.
In addition, unlike regression loss which deteriorates performance, reconstruction loss does not harm performance because it induces learning the semantic relationship than direct regression supervision. 

\begin{table}[t]
\centering
\small
\caption{mADE$_6$ according to actor and lane masking ratio.}
\vspace{-5pt}
\label{tab:ablation_mask_table}
\begin{tabular}{cc|ccccc}
\toprule
\multicolumn{2}{c|}{\multirow{2}{*}{\begin{tabular}[c]{@{}c@{}}INTER $\rightarrow$ Lyft\\ (1/3/0.1)\end{tabular}}} & \multicolumn{5}{c}{Lane Masking Ratio} \\ 
\multicolumn{2}{c|}{}  & 0.1 & 0.3 & 0.5 & 0.7 & 0.9  \\ \hline
% \multicolumn{2}{c|}{\begin{tabular}[c]{@{}c@{}}mADE$_6$ @\\  INTER $\rightarrow$ Lyft\\ (1/3/0.1)\end{tabular}} & \multicolumn{5}{c}{Lane Masking Ratio} \\ 
% \multicolumn{1}{c}{} &  & 0.1 & 0.3 & 0.5 & 0.7 & 0.9  \\ \hline
\multirow{5}{*}{\begin{tabular}[c]{@{}c@{}}Actor \\ Masking \\ Ratio\end{tabular}} & 0.1 & 0.418 & 0.407 & 0.464 & 0.481 & 0.446 \\
                                                                                   & 0.3 & 0.423 & 0.443 & 0.443 & 0.445 & 0.390 \\
                                                                                   & 0.5 & 0.417 & 0.455 & 0.515 & 0.391 & 0.435 \\
                                                                                   & 0.7 & 0.567 & 0.435 & 0.448 & 0.453 & 0.491 \\
                                                                                   & 0.9 & 0.447 & 0.421 & 0.494 & 0.417 & 0.398 \\ \bottomrule
\end{tabular}

% \begin{tabular}{cc|ccccc}
% \toprule
% \multicolumn{2}{c|}{\multirow{2}{*}{\begin{tabular}[c]{@{}c@{}}mADE$_6$ @ \\ INTER $\rightarrow$ Lyft\\ (1/3/0.1)\end{tabular}} \vspace{15pt}} & \multicolumn{5}{c}{Lane masking ratio}     \\
% \multicolumn{2}{c|}{}                                                                                             & 0.1   & 0.3   & 0.5   & 0.7   & 0.9   \\ \hline
% \multirow{5}{*}{\begin{tabular}[c]{@{}c@{}}Actor \\ masking \\ Ratio\end{tabular}}             & 0.1            & 0.418 & 0.407 & 0.464 & 0.481 & 0.446 \\
%                                                                                                & 0.3            & 0.423 & 0.443 & 0.443 & 0.445 & 0.390 \\
%                                                                                                & 0.5            & 0.417 & 0.455 & 0.515 & 0.391 & 0.435 \\
%                                                                                                & 0.7            & 0.567 & 0.435 & 0.448 & 0.453 & 0.491 \\
%                                                                                                & 0.9            & 0.447 & 0.421 & 0.494 & 0.417 & 0.398 \\ \bottomrule
% \end{tabular}
\vspace{-5pt}
\end{table}

\subsection{Actor-specific token}
\begin{figure}[t]
    \centering
    \includegraphics[width=0.99\linewidth]{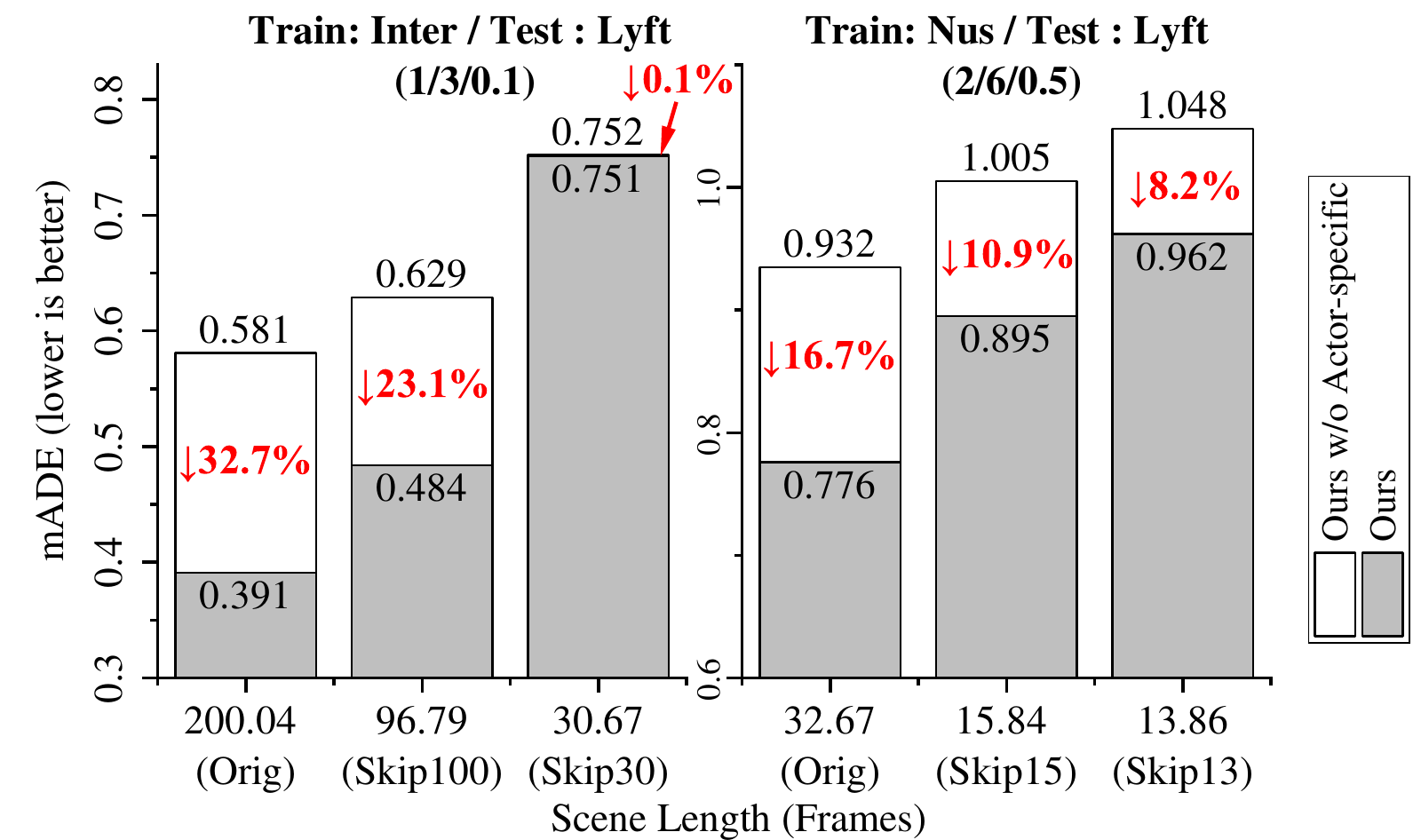}
    \vspace{-5pt}
    \caption{Effect of scenario length to the effectiveness of actor-specific token. As the scenario length shortens with manual skipping, its effectiveness diminishes because the duration available for the actor-specific token to adapt is reduced.
    In real-world applications, driving scenarios are continuous, resulting in maximal efficacy of the proposed method.}
    \label{fig:ablation_scene_skip}
    \vspace{-8pt}
\end{figure}

Table.~\ref{tab:ablation_actor_specific} presents results for the baseline without adaptation, our method without actor-specific tokens, and our full method.
The second column reveals that even without actor-specific tokens, the prediction performance is 0.581 and 0.931, surpassing MEK's 0.746 and 1.006.
However, incorporating actor-specific tokens for instance-aware adaptation yields notable improvements of 32.7\% and 16.7\% for short-term and long-term experiments, respectively.
The difference in performance between short-term and long-term is influenced by the scene length in the dataset.
The average scene length for short-term data with a 0.1 time interval is 200.04, significantly longer than the average of 32.67 for long-term data with a 0.5 time interval.
Intuitively, as scene length increases, the time spent observing previously adapted actors also increases, enhancing the effectiveness of actor-specific tokens.
To verify this, Fig.~\ref{fig:ablation_scene_skip} adjusts scene length arbitrarily by skipping to the next scene in data loading once a specific scene length is exceeded.
The results confirm that as scene length decreases by skipping scenes earlier, the effectiveness of actor-specific tokens diminishes in both short-term and long-term scenarios.

\begin{table}[t]
\footnotesize
\centering
\caption{Effect of actor-specific token in mADE$_6$/mFDE$_6$.
The proposed method enhances adaptation performance by learning actor-wise motion characteristics.}
\vspace{-5pt}
\label{tab:ablation_actor_specific}
\begin{tabular}{l|ccc}
\toprule
                                                                 Exp. & Baseline      & \begin{tabular}[c]{@{}c@{}}Ours w/o \\ Actor-specific\end{tabular} & Ours (Full)   \\ \hline
\begin{tabular}[c]{@{}l@{}}INTER → Lyft \\ (1/3/0.1)\end{tabular} & 1.553 / 3.458 & 0.581 / 1.151                                                             & \textbf{0.391} / \textbf{0.824} \\ \hline
\begin{tabular}[c]{@{}l@{}}nuS → Lyft \\ (2/6/0.5)\end{tabular}   & 1.108 / 2.597 & 0.932 / 2.220                                                             & \textbf{0.776} / \textbf{1.820} \\ \bottomrule
\end{tabular}
\end{table}

%% file: sec/7_conclusion.tex
\section{Conclusion}
We propose a test-time training method for trajectory prediction by incorporating the MAE and actor-specific token memory. 
The introduced MAE objective addresses a limitation of conventional online learning, preventing the loss of representations learned from source data.
Consequently, our approach enables learning deeper layers, leading to improved representations and enhanced predictions even for out-of-distribution samples.
The integration of actor-specific tokens during test-time allows for instance-wise learning of motion patterns, resulting in substantial performance improvements. 
This approach, particularly effective in continuous real-world autonomous driving scenarios without scene breaks, demonstrates significant efficacy and holds promise for practical applications.

%% file: sec/X_suppl_arxiv.tex
\clearpage
\setcounter{page}{1}
\maketitlesupplementary

\section{Metric definitions}
\paragraph{minimum Average Displacement Error (mADE)}
The ADE measures the average L2 distance between the predicted trajectory $\hat{\textbf{x}}_t^n=(x_{t-t_h:t+t_f}^n, y_{t-t_h:t+t_f}^n)$ and its corresponding ground truth $\textbf{x}_t^n$ for $n$-th agent and $t$-th time step.
The $\text{mADE}_k$ represents the minimum ADE over the $k$ most likely predictions, and is found for all scenes \textbf{S} in the test set.
\begin{equation}
    \text{ADE} = \frac{1}{|\textbf{S}|}\sum_{s=0}^{|\textbf{S}|}\frac{1}{N}\sum_{n=0}^N{\frac{1}{T_{s}}\sum_{t=0}^{T_s}{\|\textbf{x}_t^n-\hat{\textbf{x}}_t^n\|_2}}
\end{equation}
\begin{equation}
    \text{mADE}_k = \min_k(\text{ADE}_{(1)}, ..., \text{ADE}_{(k)})
\end{equation}

\paragraph{minimum Final Displacement Error (mFDE)}
The FDE measures the L2 distances between the predicted final point $\hat{\textup{x}}_{T_s}^n = (x_{T_s}^n, y_{T_s}^n)$ of the prediction and ground truth.
The $\text{mFDE}_k$ represents the minimum FDE over the $k$ most likely predictions, and is found for all scenes \textbf{S} in the test set.
\begin{equation}
    \text{FDE} = \frac{1}{|\textbf{S}|}\sum_{s=0}^{|\textbf{S}|}\frac{1}{N}\sum_{n=0}^N{\|\textup{x}_{T_s}^n-\hat{\textup{x}}_{T_s}^n\|_2}
\end{equation}
\begin{equation}
    \text{mFDE}_k = \min_k(\text{FDE}_{(1)}, ..., \text{FDE}_{(k)})
\end{equation}

\paragraph{Miss Rate (MR)}
% cite nuScenes
MR is the proportion of missed predictions over all predictions.
Following the nuScenes dataset, we defined the prediction whose maximum pointwise L2 distance to ground truth is greater than 2 meters as missed predictions.
$\text{MR}_k$ take the $k$ most likely predictions and determine whether they are missed predictions or not.
If there are $m$ misses over total $n$ predictions, MR would be $\frac{m}{n}$.

\section{Model details}
We employ the same model architecture as our backbone model, ForecastMAE.
Actor-specific tokens, represented by $\bar{\alpha} \in \mathbb{R}^{C \times D}$, comprise learnable parameters for each class ($C$).
\begin{equation}
    \bar{\alpha} \in \mathbb{R}^{C \times D} : \left\{ \alpha(c) \in \mathbb{R}^{D} \right\}^C
\end{equation}
Each $n^{th}$ actor corresponds to an actor class token $\alpha_n(c)$ within a specific class, where during offline training, all actors within the same class share the same class token $\alpha(c)$ regardless of individual instances.
During test-time training and online evaluation, leveraging historical motion patterns for each actor instance at a specific time ($t$), we refine the actor-specific token $\alpha_n^t(c)$ to capture distinct actor-specific motion using Algorithm~\ref{alg:actor_specific_memory}.

In our MAE training approach, we implement random masking for lane and complementary masking for actors. 
Random lane masking involves replacing a segment of lane embeddings with a learnable lane masking token. 
Complementary actor masking entails randomly selecting a subset of actors, replacing their future trajectory embeddings with a learnable trajectory masking token. 
Other actors have their past trajectory embeddings replaced with the same learnable trajectory masking token.

\section{Training details}
As our focus is set on test-time training for trajectory prediction, baseline comparisons are made with existing \textit{test-time adaptation methods}. 

\paragraph{DUA.} Retaining the hyperparameters from the official implementation, we use \textit{pre\_momentum} ($\rho_0$) = 0.1, \textit{decay\_factor} ($w$) = 0.94, and \textit{min\_momentum\_constant} ($\zeta$) = 0.005 for momentum updates.

\paragraph{TENT w/ sup.} Employing identical hyperparameters, including learning rate and weight decays, we update layers of BatchNorm1d, BatchNorm2d, BatchNorm3d, SyncBatchNorm, and LayerNorm types within the backbone model.

\paragraph{MEK.} Adhering to the online learning methodology from the original paper and importing the MEK optimizer from \textit{MEKF-MAME}. 
In the absence of multi-modal prediction loss mention, we utilize the WTA loss, commonly employed for multi-modal trajectory prediction.

\paragraph{AML.} Employing the hyperparameters and methodology from the official implementation for the nuS $\rightarrow$ Lyft experiment, adapted to our backbone. We set $\alpha_{init}$ = 0.0001, learning\_rate = 0.001, and sigma\_eps$_{init}$ = 0.102.
The regression loss of AML is different from ours and MEK.
While ours and MEK use delayed historical and future trajectories ($\mathcal{X}_{t-\tau}, \mathcal{Y}_{t-\tau}$), AML uses current historical trajectory ($\mathcal{X}_{t}$) for compute regression loss for meta learning.
It has advantage in using current motion, but disadvantage in not using full historical and future trajectories. 

\paragraph{Ours}
% Unlike the original \textit{Forecast-MAE} used MAE framework for pretraining, we use it with regression loss for joint training.
We jointly train regression and reconstruction losses with equal weightage (set as 1). 
The reconstruction loss encompasses history, future, and lane reconstructions with weights of 1, 1, and 0.35, respectively. 
We update all types of layers across all depths.

\section{Algorithm of actor-specific token memory}
\label{sec:actor_specific_algorithm}
The actor-specific token is implemented as a learnable embedding of a transformer, and stored in a memory dictionary where actor instance ID/corresponding tokens are key/values; as each token is a 128-dim vector, countless actors can be stored in memory.

While we know when an actor disappears and reappears during offline training, this is unavailable during online training/inference. 
% Ideally an object tracking network will be used to re-id actor instance IDs during online training/inference. 
% used as keys are obtained from a separate object tracking network, our framework can become dependent. However,  
How the token memory evolves is closely related to how an object-tracking network tracks actor IDs during online training and inference.
% During online training/inference, tracking actor IDs follows tracking protocol of an object tracking network. 
During online training/inference, actor IDs are tracked by an object-tracking network. 
% The evolving and initialization protocol of token memory is closely related to the tracking protocol.
In the case where an actor disappears/reappears 
and the tracker succeeds in restoring the actor ID, the ID is used to retrieve the corresponding actor-specific token from the dictionary. 
However, when the tracker fails and assigns a new ID, our method also re-initializes the token.
% from the averaged class-embedding ($\bar{\alpha}_{scene}$) in line 268 of the main paper. 
% However, when the tracking network fails and assigns a new ID, our method initializing actor tokens from a class average helps ensure a robust re-initialization. 
If an actor appears, an object tracker has its own strategy for actor-instance initialization.
As the proposed protocol can share the strategy of the tracker, our method can be integrated into the perception system seamlessly.
% For example, when an actor disappears, we maintain actor-specific token if tracker maintain its ID and vice versa.

The comprehensive algorithm delineating the actor-specific memory is outlined in Algorithm~\ref{alg:actor_specific_memory}.
This specialized memory repository encompasses individual actor-specific tokens, dynamically evolving as scene-relative time passes.
Upon a scene transition, the actor-specific tokens are collected by class, averaged, and subsequently passed on to the succeeding scene.

\begin{algorithm}
\caption{Actor-Specific Token Memory}
\label{alg:actor_specific_memory}
\textbf{Given} $s$ = the scene index of a set of scenes \textbf{S}, \\
$n$ = the $n^{th}$ actor seen in the scene,  \\
$c$ = the class index from a set of $C$ classes, \\
$\alpha$ = actor specific tokens, \\
$t$ = scene relative time, \\
$T_s$ = time length for scene $s$, \\
$(c)$ = the class type of actor, \\
$N_c$ = the number of actors of class $c$\\
$E$ = the network Encoder\\
$D$ = the network Decode, and \\
$\tau$ = the delayed time stamp:
    \begin{algorithmic}[1]
    % \Require $\bar{\alpha}_\textbf{train}$
    % \Ensure $y = x^n$
    % \Initialize 
    \State $\bar{\alpha}_\textbf{scene(0)}(c) \gets \bar{\alpha}_\textbf{train}(c), \; \forall c \in C$
    \For{$s=0:|\textbf{S}|$}
    \State $n \gets 0$
    \State $\boldsymbol{\mathcal{A}} \gets \{\}$ \Comment{This serves as the memory bank}
    \For{$t=0:T_s$}
    \For{$\alpha_{new}$} \Comment{$\forall$ new actor $\alpha_{new}$ at time t}
    \State $\alpha_{n}^{t}(c) \gets \bar{\alpha}_{\textbf{scene}(s)}(c)$ 
    \State $\boldsymbol{\mathcal{A}}.insert(\alpha_{n}^{t}(c))$
    \State $n \gets n+1$
    \EndFor
    \If {$t\bmod{\tau} \equiv 0$} \Comment{Train every $\tau$ steps}
    \If{$\alpha^{t}_{n}(c) \in {\textbf{scene}(s)}_t$}
    \State $\alpha^{t+1}_n(c) \gets \alpha^{t}_n(c) \text{-} \frac{\partial \mathbb{E}^{t-\tau}(\alpha_{n}(c), \mathcal{X}, \mathcal{Y}, \mathcal{M})}{\partial \alpha^{t-\tau}_{n}}$
    \State \qquad \qquad $\forall \; \alpha^{t}_{n}(c) \in \boldsymbol{\mathcal{A}}$  
    \Else
    \State $\alpha^{t+1}_n(c) \gets \alpha^{t}_n(c)$
    \State \qquad \qquad $\forall \; \alpha^{t}_{n}(c) \in \boldsymbol{\mathcal{A}}$  
    \EndIf
    % \State $\alpha^{t+1}_n(c) \gets
    %     \begin{cases}
    %     \alpha^{t}_n(c), \textbf{if }\alpha^{t}_{n}(c) \notin \textbf{scene(s)}_t \\
    %     \alpha^{t}_n(c) - \frac{\partial \mathbb{E}(\alpha^{t}_{n}(c), \mathcal{X}, \mathcal{Y}, \mathcal{M})}{\partial \alpha^{t}_{n}}, \textbf{else}
    %     \end{cases}
    %     $\;
    % \gets (\alpha^{t} - \frac{\partial \mathbb{E}(\boldsymbol{\mathcal{A}}, \mathcal{X}, \mathcal{Y}, \mathcal{M})}{\partial \boldsymbol{\mathcal{A}}}),$
    \EndIf
    \State $\mathcal{Y}_{t} = D(E(\mathcal{X}_{t}, \mathcal{M}_{t}, \boldsymbol{\mathcal{A}}))$ \Comment{Online-Eval}
    \EndFor
    \For{$c=0:|C|$}
    \State $\bar{\alpha}_\textbf{scene(s+1)}(c) \gets \frac{1}{N_c} \sum_n^{N_c} \alpha^{T_s}_{n}(c)$
    \EndFor
    
    \EndFor
    \end{algorithmic}
\end{algorithm}

\section{Datasets}
We construct four different datasets in the same format using \textit{trajdata}, a unified framework for trajectory prediction data.
Our preprocessing method aligns closely with the data preprocessing approach of \textit{Forecast-MAE}, with a couple of modifications from the original methodology.

Primarily, we omit certain information used in the original method, such as \textit{is\_intersections} and \textit{lane\_attribute}, due to their non-universal availability across all datasets. 
Additionally, our lane parsing method diverges in specifics. 
We focus on lane information within a 50-meter radius of ego-agents, interpolating each lane centerline to standardize point distances within the lane to 1 meter. 
Furthermore, individual lanes are divided into segments, each segment limited to a maximum length of 20 meters.

The parsing of data within a scenario, facilitated by the scenario-based parsing protocol of \textit{trajdata}, hinges on parsing time configuration and absolute scenario length. 
Our method utilizes two distinct time configurations for prediction purposes: a 0.9-second past (including the current second) and a 3.0-second future interval with a 0.1-second time interval for short-term prediction. 
Conversely, for long-term prediction, we opt for a 2.0-second past (including the current second) and a 6.0-second future interval with a 0.5-second time interval. 
This configuration results in shorter time intervals yielding longer scenario lengths for short-term prediction compared to longer time intervals for long-term prediction.

Tables~\ref{tab:supp_dataset_long} and \ref{tab:supp_dataset_short} display the mean scenario lengths across datasets for long-term and short-term prediction, respectively. 
Our update step ($\tau$) is set as $t_f$, ensuring that the target data encompasses a scenario length exceeding $t_f$ (12 for long-term and 30 for short-term). 
Consequently, for long-term prediction, nuS and Lyft datasets are designated as target datasets, while for short-term prediction, nuS, Lyft, and Way datasets serve as the target datasets. 
Please note that the real-world application is a continuous setting without scene transition, so setting the target dataset as a sufficiently long scenario length is a realistic experiment setting.

\begin{table}[h]
    \centering
    \caption{Mean scenario length of each \textit{val} dataset in long-term prediction configurtaion.}
    \label{tab:supp_dataset_long}
    \begin{tabular}{l|ccc} \toprule
    \begin{tabular}[c]{@{}l@{}}Long-term\\ (2/6/0.5)\end{tabular} & nuS & Lyft & Way \\ \hline
    scene length                                                  & 23.0  & 32.7 & 2.0  \\ \bottomrule
    \end{tabular}
\end{table}
\begin{table}[h]
\centering
\caption{Mean scenario length of each \textit{val} dataset in short-term prediction configurtaion.}
\label{tab:supp_dataset_short}
\begin{tabular}{l|cccc} \toprule
\begin{tabular}[c]{@{}l@{}}Short-term\\ (1/3/0.1)\end{tabular} & nuS   & Lyft  & Way  & INTER \\ \hline
scene length                                                   & 150.4 & 200.0 & 50.5 & 6.3   \\ \bottomrule
\end{tabular}
\end{table}

\section{Further quantitative results}
\subsection{Additional cross-dataset adaptation results}
We conducted exhaustive cross-dataset adaptation experiments for both long-term and short-term predictions, detailed in Tab.\ref{tab:supp_dataset_long} and Tab.\ref{tab:supp_dataset_short}, respectively. 
These results notably demonstrate the distinct superiority of our method over existing Test-Time Adaptation (TTA) and online learning methodologies.

\subsection{More metrics comparison}
mADE$_1$ and Missrate results in Tab.~\ref{tab:metrics} show ours is still effective.
Missrate (MR) is widely-used and measures precision.
\begin{table}[h]
\caption{Comparison with other metrics (lower is better)}
\label{tab:metrics}
\resizebox{\linewidth}{!}{
% \scriptsize
\begin{tabular}{c|cccc}
\hline
\rowcolor[HTML]{EFEFEF} 
\begin{tabular}[c]{@{}c@{}} (mADE$_1$/Missrate)\end{tabular}                           & Source only   & TENT          & MEK           & Ours                   \\ \hline
\cellcolor[HTML]{EFEFEF}\begin{tabular}[c]{@{}c@{}}INTER → nuS (1/3/0.1)\end{tabular} & 3.030 / 0.338 & 1.726 / 0.336 & 1.461 / 0.291 & \textbf{1.239/ 0.154}  \\
\cellcolor[HTML]{EFEFEF}\begin{tabular}[c]{@{}c@{}}nuS → Lyft (2/6/0.5)\end{tabular}  & 2.546 / 0.362 & 2.473 / 0.346 & 2.412 / 0.323 & \textbf{1.918 / 0.227} \\ \hline
\end{tabular}
}
\end{table}

\subsection{More ablation of actor-specific token on other datasets}
We included only two datasets due to page limits; experiments on other datasets (Lyft, Way) in Tab.~\ref{tab:additional_dataset} still show that our method is effective.
\begin{table}[h]
\vspace{-10pt}
\caption{Effect of actor-specific memory on additional datasets}
\label{tab:additional_dataset}
\resizebox{\linewidth}{!}{
\begin{tabular}{l|ccc}
\hline
\rowcolor[HTML]{EFEFEF} 
Exp (mADE$_6$/mFDE$_6$)                               & Baseline      & Ours w/o A-s  & Ours                   \\ \hline
\cellcolor[HTML]{EFEFEF}Lyft → nuS (1/3/0.1) & 0.506 / 1.102 & 0.420 / 0.934 & \textbf{0.357 / 0.770} \\
\cellcolor[HTML]{EFEFEF}Way → Lyft (2/6/0.5) & 0.638 / 1.404 & 0.594 / 1.311 & \textbf{0.549 / 1.171} \\ \hline
\end{tabular}
}
\end{table}

\subsection{Efficiency under long-term configuration}
\begin{figure}[h]
    \centering
    \includegraphics[width=\linewidth]{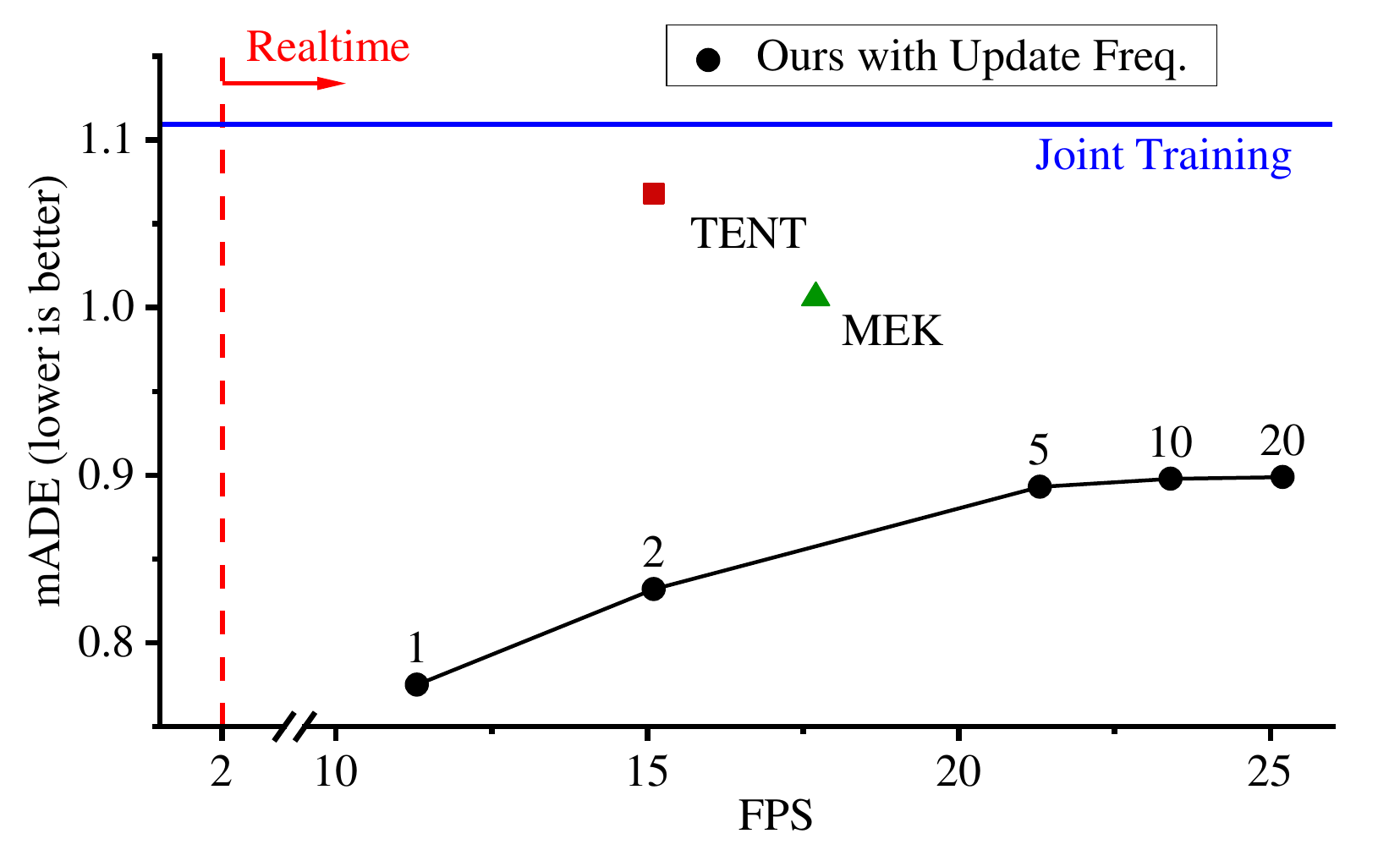}
    \caption{mADE$_6$ and FPS of our method and the baselines in nuS $\rightarrow$ Lyft long-term experiment (2/6/0.5). }
    \label{fig:efficiency_long}
\end{figure}

In Fig.~\ref{fig:efficiency_long}, we present another efficiency comparison experiment between our method and baseline approaches in the long-term configuration. Demonstrating superiority in both accuracy and efficiency, our method outperforms the baseline methods in this setting.
All methods achieve real-time execution, surpassing the 2FPS benchmark set by the 0.5 time interval of data acquisition. However, in the long-term configuration with an adaptation step of 5, MEK shows a 17.7 improvement in computational efficiency but experiences a significant decline in accuracy.
TENT's efficiency remains nearly consistent with the short-term configuration, but its accuracy doesn't exhibit significant improvement from joint training, making it noteworthy in this context.

\section{Expanded visual results}
\subsection{Comparative analysis with baseline methods}
\begin{figure*}[h]
    \centering
    \includegraphics[width=0.9\linewidth]{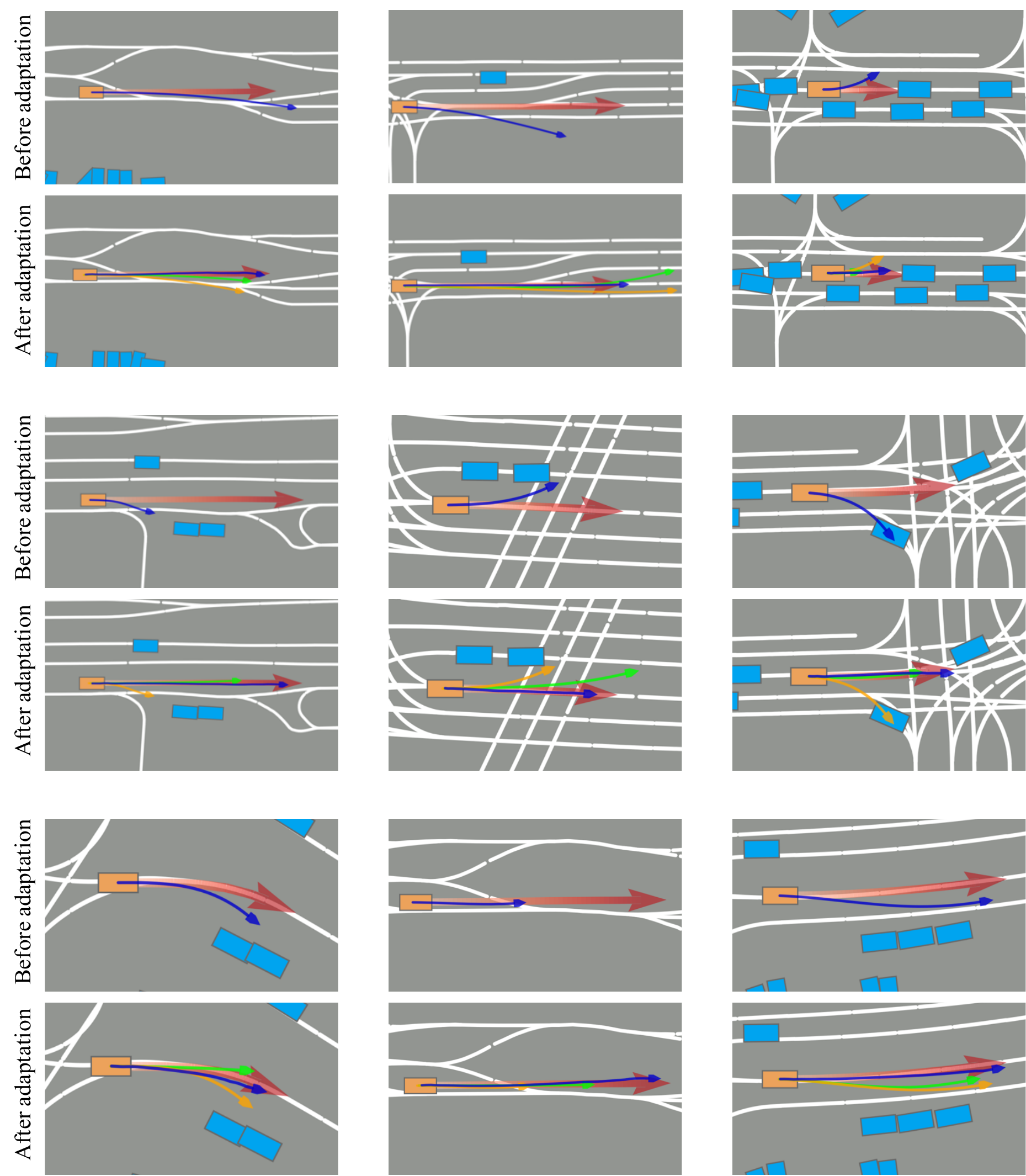}
    \caption{Additional visual comparison of adaptation results of our method against the baselines. For each image pair, the above shows prediction results before adaptation, and the below shows prediction results after adaptation via ours (blue arrow), TENT w/ sup (orange arrow), and MEK (green arrow). Red arrows denote GT future trajectories. Sky blue and orange boxes refer to surrounding actors and actors to be predicted. We depicted only one actor result and one mode among multi-modal predictions closest to the GT for visual simplicity. Please note that our method is multi-modal prediction for all actors method. }
    \label{fig:results_compraison_suppl}
\end{figure*}

Figure.~\ref{fig:results_compraison_suppl} showcases additional qualitative comparison results between our method and the baseline methods. 
Notably, TENT w/ sup exhibits minimal adaptation in comparison to the other methodologies.
Regarding MEK, the supervision signal tends to lead to underfitting or over-adaptation, primarily adjusting the last layer weights of the decoder without inducing the model to acquire robust representations.
Conversely, our method demonstrates stable adaptability, particularly excelling in challenging scenarios.

\subsection{Reconstruction results}
\begin{figure*}[h]
    \centering
    \includegraphics[width=\linewidth]{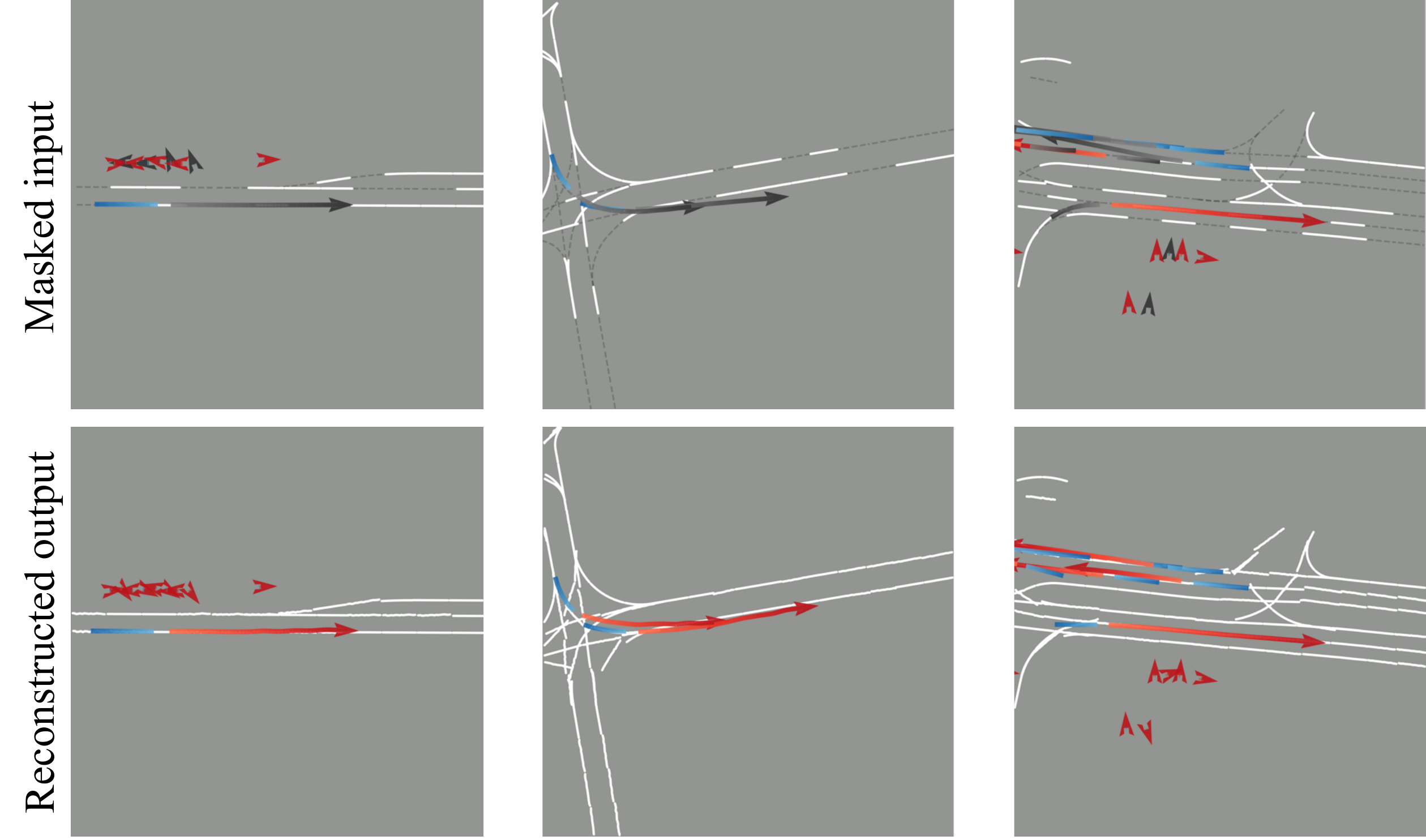}
    \caption{The first row indicates masked samples, and the row below shows the reconstructed outputs. The blue/red arrows indicate
historical/future trajectories. The black arrows refer to the masked
trajectories. The white lines are the lane centerlines, and the gray
dashed lines are the masked lane centerlines.}
    \label{fig:recon_suppl}
\end{figure*}

In Fig.~\ref{fig:recon_suppl}, additional reconstruction results are showcased. 
These reconstruction examples are trained offline using the source dataset and subsequently adapted during test time using the target dataset via reconstruction loss.
Upon examination of the three reconstruction instances, it's evident that the method appropriately learns lane structures, emphasizing an understanding of interactions among lane segments and actor trajectories. 
Furthermore, in the second column, the proper reconstruction of future trajectories for actors is observed, showcasing an understanding of both lane structure and the motion of following/leading actors.

\subsection{Extended multi-modal prediction findings}
\begin{figure*}[t]
    \centering
    \includegraphics[width=\linewidth]{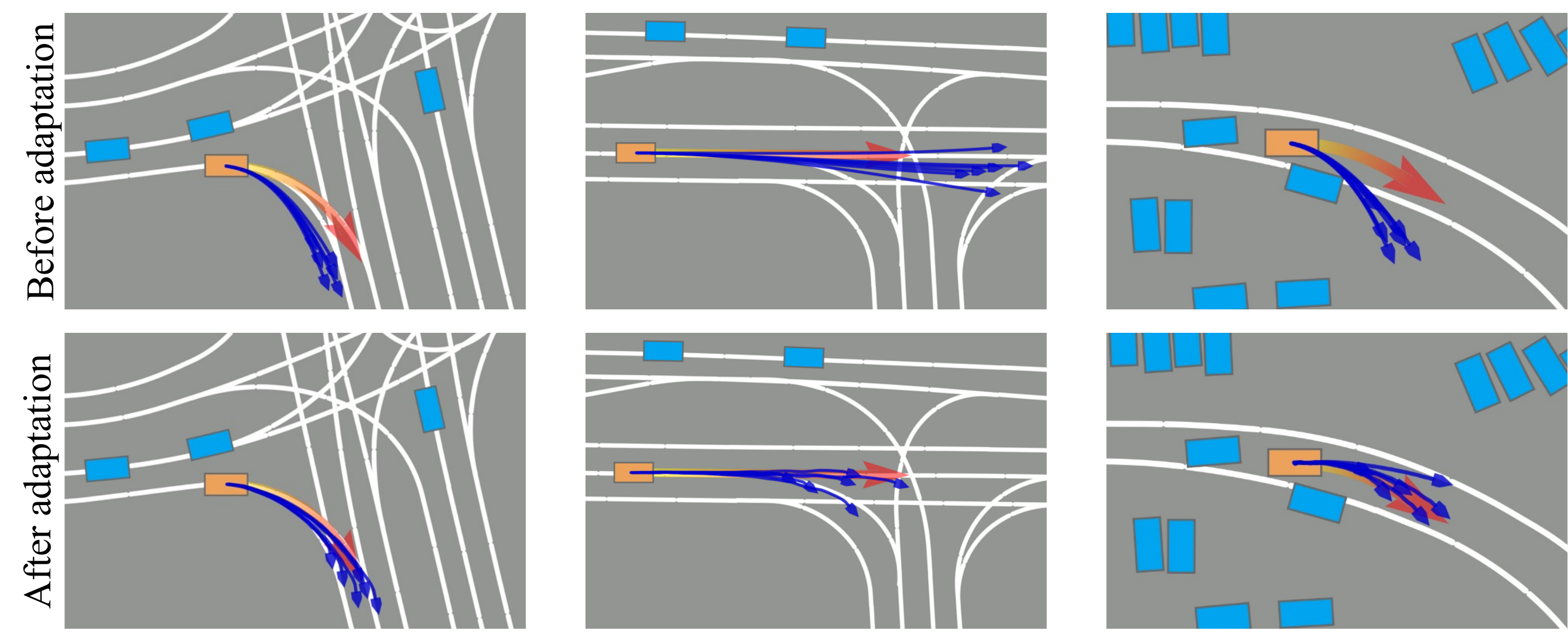}
    \caption{Multi modal prediction results (blue arrow) before and
after adaptation via our method. Ours generates elaborate samples
that consider interaction between lane or other actor due to representation learning, which cannot be learned from
the GT (red arrow) using regression loss.
}
    \label{fig:multimodal_suppl}
\end{figure*}

Figure.~\ref{fig:multimodal_suppl} presents additional multi-modal prediction results obtained through our adaptation method. 
In the first and second columns, our approach generates diverse and plausible prediction samples while comprehending road structures adeptly. 
Notably, it showcases an understanding of complex road structures, such as turn scenarios.
Moving to the third column, our method also demonstrates an understanding of interactions with surrounding actors. 
It illustrates instances where our model generates trajectories that avoid collisions by either surpassing nearby actors or transitioning to another lane. 
In contrast, the non-updated version produces trajectories that could lead to collisions with the front right actor.

\clearpage

% \afterpage{
\begin{sidewaystable}[h]
\centering
\caption{All cross-dataset adaptation experiments in long-term configuration. We use all nuS, Lyft, Way for the source dataset, and use nuS, Lyft for the target datasets, ensuring they exceed the adaptation step of 12.}
\label{tab:my-table}
% \begin{adjustbox}{angle=270}
\resizebox{0.6\textwidth}{!}{
\setlength\extrarowheight{5pt}
% \resizebox{0.6\textwidth}{!}{
\begin{tabular}{l|ccccc} \toprule
\multicolumn{1}{c|}{\multirow{2}{*}{\begin{tabular}[c]{@{}c@{}}mADE$_6$ \\ / mFDE$_6$\end{tabular}}} & \multicolumn{5}{c}{Long-term exp (2/6/0.5)}                                                                                  \\ \cline{2-6} 
\multicolumn{1}{c|}{}                                                                          & nuS → Lyft & Way → Lyft & Way → nuS & \multicolumn{1}{c|}{Lyft → nuS} & Mean \\ \hline
Source Only                                                                                    & 1.122 / 2.577          & 0.621 / 1.347          & 1.153 / 2.220         & \multicolumn{1}{c|}{1.434 / 3.178}          &   1.083 / 2.331   \\
Joint Training                                                                                 & 1.108 / 2.597          & 0.638 / 1.404          & 1.091 / 2.031         & \multicolumn{1}{c|}{1.398 / 3.109}          &   1.059 / 2.285   \\
DUA                                                                                            & 1.365 / 3.257          & 0.790 / 1.868          & 1.270 / 2.634         & \multicolumn{1}{c|}{1.585 / 3.607}          &   1.253 / 2.842   \\
TENT (w/ sup)                                                                                    & 1.068 / 2.514          & 0.628 / 1.381          & \underline{1.077 / 2.012}         & \multicolumn{1}{c|}{\underline{1.395 / 3.102}}          &  1.042 / 2.252    \\
MEK ($\tau=t_f/2$)                                                               & 1.079 / 2.597          & 0.629 / 1.404          & 1.079 / 2.031         & \multicolumn{1}{c|}{1.396 / 3.109}          &  1.046 / 2.285    \\
MEK ($\tau=t_f$)                                                                 & \underline{1.006 / 2.369}          & \underline{0.615 / 1.351}          & 1.117 / 2.140         & \multicolumn{1}{c|}{1.426 / 3.119}          &  \underline{1.041 / 2.245}    \\
AML (\textit{K$_0$})                                                                                       & 1.787 / 3.067          & 1.322 / 2.571          & 1.618 / 2.999         & \multicolumn{1}{c|}{1.866 / 3.494}          &  1.648 / 3.033    \\
AML (\textit{full})                                                                                     & 1.462 / 2.573          & 0.977 / 2.184          & 1.495 / 2.978         & \multicolumn{1}{c|}{1.698 / 3.367}          &    1.408 / 2.776  \\

\rowcolor[rgb]{0.9,0.9,0.9} Ours                                                                                           & \textbf{0.776 / 1.820}          & \textbf{0.549 / 1.171}          & \textbf{0.996 / 1.784}         & \multicolumn{1}{c|}{\textbf{1.254 / 2.802}}          &   \textbf{0.891 / 1.888}   \\ \bottomrule
\end{tabular}
}
% \end{adjustbox}
\end{sidewaystable}
\begin{sidewaystable}[h]
\caption{All cross-dataset adaptation experiments in shoft-term configuration. We use all nuS, Lyft, Way, INTER for the source dataset, and use nuS, Lyft, Way for the target datasets, ensuring they exceed the adaptation step of 30.}
\label{tab:my-table}
\centering
% \begin{adjustbox}{angle=270}
\resizebox{\textwidth}{!}{
\setlength\extrarowheight{5pt}
\begin{tabular}{l|cccccccccc} \toprule
\multicolumn{1}{c|}{\multirow{2}{*}{\begin{tabular}[c]{@{}c@{}}mADE$_6$ \\ / mFDE$_6$\end{tabular}}} & \multicolumn{10}{c}{Short-term exp (1/3/0.1)}                                                                                                                                                                                           \\ \cline{2-11} 
\multicolumn{1}{c|}{}                                                                          & INTER → nuS   & INTER → Lyft   & INTER → Way & nuS → Lyft & nuS → Way & Way → Lyft & Way → nuS & Lyft → nuS & \multicolumn{1}{c|}{Lyft → Way} & Mean \\ \hline
Source Only                                                                                    & 1.047 / 2.247 & 1.391 / 2.945 & 0.458 / 1.052           & 0.233 / 0.553          & 0.431 / 1.031         & 0.191 / 0.424          & 0.382 / 0.761         & 0.484 / 1.096          & \multicolumn{1}{c|}{0.621 / 1.348}          &   0.582 / 1.273   \\
Joint Training                                                                                 & 1.116 / 2.445 & 1.553 / 3.458 & 0.398 / 1.028           & 0.233 / 0.504          & 0.472 / 1.125         & 0.185 / 0.415          & 0.374 / 0.718         & 0.506 / 1.102          & \multicolumn{1}{c|}{0.918 / 1.935}          &   0.639 / 1.414   \\
DUA                                                                                            & 1.118 / 2.455 & 1.516 / 3.352 & 0.398 / 1.031           & 0.287 / 0.724          & 0.516 / 1.294         & 0.197 / 0.433          & 0.396 / 0.781         & 0.474 / 1.118          & \multicolumn{1}{c|}{0.981 / 2.081}          &   0.654 / 1.474   \\
TENT (w/ sup)                                                                                    & 1.102 / 2.423 & 1.519 / 3.405 & \underline{0.375 / 0.988}           & 0.226 / 0.485          & 0.448 / 1.071         & 0.176 / 0.398          & \underline{0.358 / 0.695}         & 0.462 / 1.009          & \multicolumn{1}{c|}{0.894 / 1.903}          &   0.618 / 1.375   \\
MEK ($\tau=t_f/2$)                                                                                     & 1.012 / 2.445 & 1.283 / 3.458 & 0.393 / 1.028           & \underline{0.220 / 0.504}          & 0.445 / 1.125         & 0.179 / 0.415          & 0.380 / 0.718         & 0.512 / 1.102          & \multicolumn{1}{c|}{0.901 / 1.935}          &  0.592 / 1.414    \\
MEK ($\tau=t_f$)                                                                                       & \underline{0.892 / 1.952} & \underline{0.746 / 1.654} & 0.409 / 1.096           & 0.231 / 0.544          & \underline{0.405 / 1.061}         & \underline{0.168 / 0.383}          & 0.373 / 0.713         & 0.508 / 1.086          & \multicolumn{1}{c|}{0.788 / 1.681}          &  \underline{0.502 / 1.130}    \\
AML (\textit{K$_0$})                                                                                         & 2.093 / 4.697 & 2.695 / 6.677 & 1.779 / 4.450           & 1.624 / 2.139          & 1.624 / 2.139         & 0.261 / 0.493          & 0.524 / 1.059         & 0.897 / 1.646          & \multicolumn{1}{c|}{1.192 / 2.513}          &  1.410 / 2.868    \\
AML (\textit{full})                                                                                       & 1.149 / 2.550 & 1.042 / 2.616 & 0.527 / 1.419           & 0.369 / 0.781          & 0.764 / 1.791         & 0.209 / 0.425          & 0.483 / 0.962         & \underline{0.454 / 0.954}          & \multicolumn{1}{c|}{\textbf{0.414 / 0.974}}          &   0.601 / 1.386   \\
\rowcolor[rgb]{0.9,0.9,0.9} Ours                                                                                           & \textbf{0.537 / 1.137} & \textbf{0.391 / 0.824} & \textbf{0.259 / 0.613}           & \textbf{0.155 / 0.316}          & \textbf{0.336 / 0.807}         & \textbf{0.155 / 0.325}          & \textbf{0.323 / 0.656}         & \textbf{0.357 / 0.770}          & \multicolumn{1}{c|}{\underline{0.515 / 1.133}}          &   \textbf{0.336 / 0.731}  \\ \bottomrule
\end{tabular}
}
% \end{adjustbox}
\end{sidewaystable}
% }

% \pagebreak

%
